\documentclass[twoside]{article}

\usepackage[preprint]{aistats2026}

\usepackage[utf8]{inputenc} 
\usepackage[T1]{fontenc}    
\usepackage{hyperref}       
\usepackage{url}            
\usepackage{booktabs}       
\usepackage{amsfonts}       
\usepackage{nicefrac}       
\usepackage{microtype}      
\usepackage[table,xcdraw,dvipsnames]{xcolor}
\usepackage{amsmath}
\usepackage{amsthm}
\usepackage{amssymb}
\usepackage{graphicx}
\usepackage{listings}
\usepackage{makecell}

\newtheorem{theorem}{Theorem}

\usepackage[style=apa, backend=biber]{biblatex}

\DeclareNameFormat{firstauthoronly}{%
  \ifnumequal{\value{listcount}}{1}
    {\namepartfamily}
    {}%
}
\DeclareNameAlias{labelname}{firstauthoronly}

\newcommand{\lmargin}{3pt}
\newcommand{\rmargin}{1pt}
\newcommand{\tmargin}{1.5pt}
\newcommand{\bmargin}{2pt}
\definecolor{light-gray}{gray}{0.96}

\lstdefinestyle{promptstyle}{
    breaklines=true,
    backgroundcolor=\color{light-gray},
    basicstyle=\ttfamily,
    frame=single, rulecolor=\color{light-gray},
    framexleftmargin=\lmargin, xleftmargin=\lmargin,
    framexrightmargin=\rmargin, xrightmargin=\rmargin,
    framextopmargin=\tmargin,
    framexbottommargin=\bmargin,
    emphstyle=\itshape
}

\lstdefinestyle{cotstyle}{
    breaklines=true,
    breakindent=0pt,
    backgroundcolor=\color{Goldenrod!50},
    basicstyle=\ttfamily,
    frame=single, rulecolor=\color{Goldenrod},
    framexleftmargin=\lmargin, xleftmargin=\lmargin,
    framexrightmargin=\rmargin, xrightmargin=\rmargin,
    framextopmargin=\tmargin,
    framexbottommargin=\bmargin,
    emphstyle=\itshape
}

\lstdefinestyle{actionstyle}{
    breaklines=true,
    breakindent=0pt,
    backgroundcolor=\color{Thistle!50},
    basicstyle=\ttfamily,
    frame=single, rulecolor=\color{Thistle},
    framexleftmargin=\lmargin, xleftmargin=\lmargin,
    framexrightmargin=\rmargin, xrightmargin=\rmargin,
    framextopmargin=\tmargin,
    framexbottommargin=\bmargin,
    emphstyle=\itshape
}

\lstdefinestyle{talkstyle}{
    breaklines=true,
    breakindent=0pt,
    backgroundcolor=\color{cyan!50},
    basicstyle=\ttfamily,
    frame=single, rulecolor=\color{cyan},
    framexleftmargin=\lmargin, xleftmargin=\lmargin,
    framexrightmargin=\rmargin, xrightmargin=\rmargin,
    framextopmargin=\tmargin,
    framexbottommargin=\bmargin,
    emphstyle=\itshape
}

\lstdefinestyle{settingstyle}{
    breaklines=true,
    breakindent=0pt,
    backgroundcolor=\color{Green!50},
    basicstyle=\ttfamily,
    frame=single, rulecolor=\color{Green},
    framexleftmargin=\lmargin, xleftmargin=\lmargin,
    framexrightmargin=\rmargin, xrightmargin=\rmargin,
    framextopmargin=\tmargin,
    framexbottommargin=\bmargin,
    emphstyle=\itshape
}

\begin{document}

\twocolumn[

\aistatstitle{GameTalk: Training LLMs for Strategic Conversation}

\aistatsauthor{%
  Victor Conchello Vendrell
  \And
  Max Ruiz Luyten
  \And
  Mihaela van der Schaar
}
\vspace{3pt}
\aistatsaddress{
  University of Cambridge \quad\quad
  \texttt{\{vc458,mr971,mv472\}@cam.ac.uk}
}
]

\begin{abstract}
Strategic decision-making in multi-agent settings is a key challenge for large language models (LLMs), particularly when coordination and negotiation must unfold over extended conversations. While recent work has explored the use of LLMs in isolated decision tasks, little attention has been given to optimizing long-term objectives through dialogue. We introduce \textbf{GameTalk}, a framework for training LLMs to make strategic decisions via multi-turn interactions. Unlike prior work that focuses on single-turn objectives or static action prediction, we train LLMs to optimize a global objective across full conversations. We achieve this by adapting fine-tuning methods like GRPO, DPO, and STaR to incorporate reward signals that depend on the entire interaction. We evaluate this approach on a suite of increasingly complex games, designed to stress different aspects of reasoning, coordination, and opponent modeling. Our results show that GameTalk significantly outperforms untrained models, especially under reward shaping, with DPO consistently yielding the strongest gains. These findings position conversational fine-tuning as a promising path for LLMs to reason, negotiate, and act in interactive environments.
\end{abstract}

\section{INTRODUCTION}

Large language models (LLMs) excel at a wide range of language tasks, from summarization~\parencite{zhangPEGASUSPretrainingExtracted2020} to problem solving \parencite{deepseek-aiDeepSeekR1IncentivizingReasoning2025}, and we typically deploy them in \textbf{multi-turn interactions} that unfold over long conversations. Yet they are usually \textbf{optimized in single-turn, static settings}, where each response is evaluated in isolation. This creates a mismatch with many real-world applications—such as negotiation, automated email assistance, and collaborative decision-making—where success depends on sustained, goal-driven dialogue. Reinforcement learning (RL) has proven effective both for multi-turn decision-making in domains like robotics~\parencite{hafnerDreamControlLearning2020} and games~\parencite{hafnerMasteringDiverseDomains2024} and for training LLMs through methods such as RLHF~\parencite{ouyangTrainingLanguageModels2022a} or GRPO~\parencite{shaoDeepSeekMathPushingLimits2024}, making it a natural candidate to address this gap. However, its potential for training LLMs to converse strategically over multiple turns remains largely unexplored~\parencite{liSingleTurnSurveyMultiTurn2025}.

We address this gap by introducing \textbf{GameTalk}, a framework for training LLMs to make strategic decisions through multi-turn interactions. Our approach leverages simple but representative games as controlled environments where conversational success can be measured explicitly via outcomes. By adapting modern fine-tuning techniques—\textit{GRPO}~\parencite{shaoDeepSeekMathPushingLimits2024}, \textit{DPO}~\parencite{rafailovDirectPreferenceOptimization2024}, and \textit{STaR}~\parencite{zelikmanSTaRBootstrappingReasoning2022}—to this setting, we enable LLMs to learn strategies that integrate reasoning, coordination, and opponent modeling.

We also introduce three novel behavioral signals to diagnose LLM weaknesses:
Internal State Evaluation, $ISE$ (Does it predict accurately its opponent's strategy?), State-Relative Performance, $SRP$ (How effective are its actions, given its beliefs about the opponent?), and Leverage Opportunity, $LO$ (Does it influence the opponent's behavior in its behalf?). We then show that standard fine-tuning methods struggle to improve these capabilities when relying only on sparse rewards from final game outcomes. Consequently, we leverage these same signals for targeted reward shaping, enabling models to learn sophisticated conversational strategies that optimize for a global objective. Experiments across increasingly complex games show that GameTalk significantly improves performance over untrained baselines, especially when enhanced with tailored reward shaping.

Our main contributions are:

\begin{itemize}
    \item \textbf{Framework:} We introduce \textbf{GameTalk}, the first RL-based framework to fine-tune LLMs for strategic multi-turn conversation.
    \item \textbf{Methodology:} We introduce a novel methodology for conversational environments by adapting existing RL algorithms and proposing \textbf{three behavioral signals} focusing on the main aspects of strategic conversations ($ISE$, $SRP$, and $LO$). These signals dissect strategic performance and are leveraged for reward shaping to improve full-interaction outcomes.
    \item \textbf{Empirical Results:} We run experiments on three representative games: Rock-Paper-Scissors, Bertrand Competition and Size-Prize Bargaining Game, chosen to analyze different capabilities. These demonstrate \textbf{consistent improvements} over baseline models, showing the effectiveness of our approach in training LLMs for strategic dialogue.
\end{itemize}

\section{RELATED WORK}

\paragraph{AIs in game settings}

A growing body of work has explored the use of games to evaluate and train AI models and specifically LLMs, revealing both their potential and their limitations as strategic agents. Several studies~\parencite{xiaMeasuringBargainingAbilities2024, kwonAreLLMsEffective2024, duanGTBenchUncoveringStrategic2024, huangHowFarAre2025, wuShallWeTeam2024, abdelnabiCooperationCompetitionMaliciousness2025a}, evaluate LLMs in multi-agent environments to test their abilities in negotiation, cooperation, and decision-making, consistently finding that while they can imitate strategic behavior, they often struggle particularly with subjective reasoning and contextually strategic responses. Other research, also trains LLMs in these domains, such as ChessGPT~\parencite{fengChessGPTBridgingPolicy2023}, where the goal is to train LLMs to play chess, and ~\textcite{chengSelfplayingAdversarialLanguage2025}, where they train on Adversarial Taboo to improve general reasoning skills. On a different note, Cicero~\parencite{metafundamentalairesearchdiplomacyteamfair+HumanlevelPlayGame2022}, and Hypothetical Minds~\parencite{crossHypotheticalMindsScaffolding2024} improve gaming performance by integrating a pretrained LLM in an architecture with other game-specific modules. Our work, however, is distinct in its focus on understanding the dynamics of the behaviours of LLMs when using current post-training methods, and proposing how to most successfully optimize the global objective.; instead of using games to evaluate static abilities or as a mere testbed for general reasoning improvements, we aim to fine-tune LLMs to internalize new strategies and use natural language communication as a central, strategic tool to achieve long-term goals in multi-turn interactions.

\paragraph{Beyond Rational Agents in Game Theory}

Standard game theory, with its assumption of perfect rationality, often fails to adapt to real-world behavior. In games like Werewolf~\parencite{wangOptimalStrategyWerewolf2024}, the Ultimatum Game~\parencite{guthExperimentalAnalysisUltimatum1982, capraroLanguagebasedGameTheory2024}, and the Keynesian Beauty Contest~\parencite{nagelUnravelingGuessingGames1995} standard models fail to account for how players actually reason and interact, leading to conclusions of very limited application. This has led to the development of Behavioral Game Theory \parencite{camererBehavioralGameTheory2011, lopezBehavioralPsychologicalGame2022}, which addresses these shortcomings by either modeling human non-optimalities as a learning process~\parencite{camererCognitiveHierarchyModel2004} (e.g., using level-$k$ reasoning \parencite{stahlEvolutionSmartnPlayers1993}) or by modifying utility functions to incorporate social preferences such as fairness and altruism \parencite{buhren2023social}. Following this line of work, we introduce a theoretical framework that unifies existing perspectives from Behavioral Game Theory at a high level in Appendix \ref{app:ModellingPlayers}, which provides concrete signals to assess the development of key behavioral traits throughout the training process.

\section{METHOD}

\subsection{GameTalk}

In this section, we present the GameTalk framework to train and analyze learning of LLMs in multi-turn settings.

\begin{figure}
    \centering
    \includegraphics[width=0.85\linewidth]{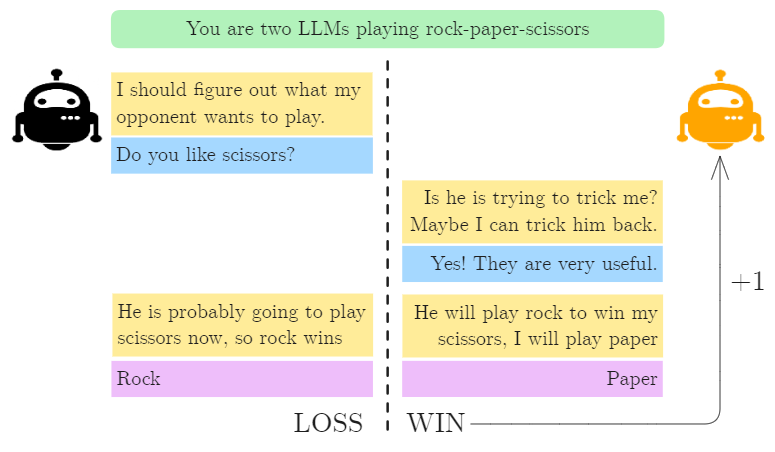}
    \caption{Example of a game of rock-paper-scissors. All texts have been shortened. The game starts with a {\color{Green} setting prompt}. The two agents first engage in a conversation, using the {\color{cyan} communication} action. When the conversation is over, they do the {\color{Thistle} game action}. That concludes the game. Before each of these actions, they use the {\color{Dandelion} private CoT}, to decide their actions, this is not shown to the opposite LLM. At the end of the game, one LLM is trained using the reward obtained from this episode.}
    \label{fig:example-game}
\end{figure}

Each interaction begins with a {\color{Green} setting prompt} with the game rules and any private information to the players. The agents then converse in turns, where they have to generate a {\color{Dandelion} Private Chain of Thought (CoT)} before each action, and then either send a {\color{cyan} message} or execute {\color{Thistle} game actions}. The private CoT promotes complex reasoning while providing transparency into the agent’s strategy. To manage the conversational flow, we utilize standard system, user, and assistant roles for prompts, opponent messages, and agent responses, and structured using HTML-like tags ({\color{Dandelion}<think>}, {\color{cyan}<talk>}, {\color{Thistle}<play>}) to clearly distinguish each part of the agents' outputs. In some games, additional information about events that have occurred during the conversation may be provided to the agents through new {\color{Green} setting prompts} introduced mid-conversation. Appendix~\ref{app:Prompts} contains all prompts used for the different games, which will be described in detail in Section~\ref{sec:mainExperiment}, including both the initial and mid-conversation prompts. Figure~\ref{fig:example-game} presents an example of a game episode.

To generate scalable training data, we train one LLM agent against a fixed LLM opponent. This setup encourages the agent to learn adaptive strategies against imperfect counterparts rather than converging toward unrealistic strategies.  Learning is driven by a single reward signal defined as the game's inherent utility $u_i$, calculated at the end of an episode from the joint actions of both players. In one-shot games, the conversation will end after one game action of each player, and in iterative games it will end after a certain number of iterations. This global reward is used to optimize all generated content, including the CoT, dialogue, and actions.

\subsection{Behavioral Signals for Strategic Analysis} \label{sec:behavioralMetrics}

To dissect the agent's strategic capabilities, we introduce three behavioral signals derived from our Deficiency-Aware Strategies framework (see Appendix \ref{app:ModellingPlayers} for details). These signals can be computed at any point during an interaction to track the evolution of the agent's strategy. To compute these signals, we approximate the opponent's true policy $\pi_{true}$ and the agent's belief about it $\pi_{belief}$ by prompting the fixed and trained LLMs, respectively, to output action probabilities from the current conversational state, as illustrated in Figure~\ref{fig:internalState}. 

These signals are motivated by the understanding that successful strategic conversation requires mastering three distinct capabilities. An agent must \textbf{(1)} form an accurate internal model to predict an opponent's actions based on the conversational history. Given this model, the agent must \textbf{(2)} select the actions that maximize its expected utility against its beliefs about the opponent. And finally, a sophisticated agent should \textbf{(3)} use dialogue to actively steer the opponent toward outcomes more favorable to them. Therefore, we designed a dedicated metric to isolate and measure each of these key skills:

\begin{figure}[!b]
    \centering
    \includegraphics[width=0.85\linewidth]{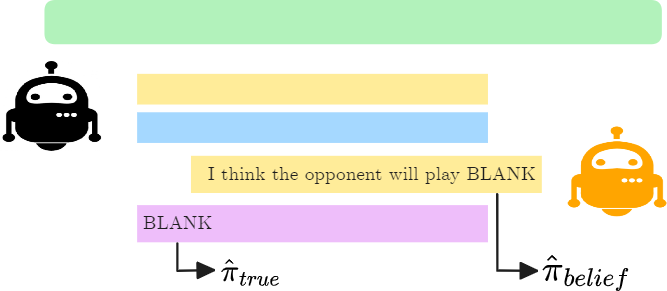}
    \caption{Illustration of how $\hat\pi_{true}$ and $\hat \pi_{belief}$ are obtained, in order to use them in \textit{ISE}, \textit{SRP} and \textit{LO}. Note that they are approximations, since they have to be computed without the {\color{Dandelion} Private CoT}, which would influence the final action.}
    \label{fig:internalState}
\end{figure}

\begin{enumerate}
    \item
\textbf{Internal State Evaluation (ISE):} Does it predict accurately its opponent's strategy?
\begin{equation} \label{eq:ise}
ISE = -D_{KL}\left(\hat \pi_{true} || \hat \pi_{belief}\right)
\end{equation}

    \item 
\textbf{State-Relative Performance (SRP):} How effective are its actions, given its beliefs about the opponent?
\begin{equation} \hspace{-10pt}
SRP = \frac{\mathbb{E}_{a_i \sim \hat \pi_{true}^u,\; a_j \sim \hat \pi_{belief}} \left[ u_i(a_i, a_j) \right]
 - U_{\text{min}}}{U_{\text{max}} - U_{\text{min}}}
\end{equation}
where $\pi_{true}^u$ represents the true policy of the agent, not the opponent, and $U_{\text{max}}$ and $U_{\text{min}}$ are the maximal and minimal expected utilities obtainable by any policy against $\hat \pi_{belief}$.

    \item 
\textbf{Leverage Opportunity (LO):} Does it influence the opponent's behavior in its behalf?
\begin{equation} \label{eq:lo}
LO = \max_\pi \mathbb{E}_{a_i \sim \pi,\; a_j \sim \hat \pi_{true}} \left[ u_i(a_i, a_j) \right]
\end{equation}

\end{enumerate}

For all of them, higher values of the metric indicate better strategic capacities.

These individual metrics are not just diagnostic; they are fundamentally linked to the agent's actual performance. The following theorem formally establishes this connection by bounding the agent's true expected utility with the three behavioral signals.

\begin{theorem}[Utility Bounds from Behavioral Signals] \label{theorem:behavioralSignals}
Let $E_{true}[u_i]$ be the agent's true expected utility, determined by its policy and its opponent's true policy. If the behavioral metrics $ISE$, $SRP$, and $LO$ are computed using the exact opponent policy $\pi_{true}$ and agent belief $\pi_{belief}$, then the utility is bounded as follows:
\begin{equation}
\begin{split}
\left[SRP \cdot (U_{\text{max}} - U_{\text{min}}) + U_{\text{min}}\right] - C \cdot \sqrt{-\frac{1}{2}ISE} 
\\
\le E_{true}[u_i] \le LO
\end{split}
\end{equation}
where $C$ is a constant representing the maximum range of the utility function $u_i$.
\end{theorem}

The proof for Theorem \ref{theorem:behavioralSignals} is provided in Appendix \ref{app:ModellingPlayers}. This relationship formalizes the intuition that achieving high utility requires an agent to first create a favorable situation (high $LO$), then understand it accurately (high $ISE$), and finally, act optimally within it (high $SRP$).

\subsection{Adaptation of RL alogrithms}

RL methods are central to our work, as we adapt them to fine-tune LLMs for multi-turn strategic interactions. The first method to show the usefulness of RL in this domain was Reinforcement Learning from Human Feedback~\parencite{ouyangTrainingLanguageModels2022}, and recent work~\parencite{deepseek-aiDeepSeekR1IncentivizingReasoning2025} has shown its huge potential for reasoning tasks, using Verifiable Rewards. To train agents within the GameTalk framework, we adapt three state-of-the-art fine-tuning algorithms to our multi-turn conversational setting:

\paragraph{Group Relative Policy Optimization ~\parencite{shaoDeepSeekMathPushingLimits2024}}
In GRPO multiple responses $\{y_i\}_{i \in G}$ for the same question $x$ are generated and then compared to train the LLM with the PPO-style policy loss
\begin{align} \label{eq:GRPO}
\begin{split}
& \mathcal{L}_{\mathrm{GRPO}}(\theta) = \mathbb{E}_{x, \{y_i\}} \left[ \frac{1}{G} \sum_{i=1}^G \min\left( \frac{\pi_\theta(y_i|x)}{\pi_{\mathrm{old}}(y_i|x)} A_i,
\right.\right. \\ & \left.\left.
\mathrm{clip}\left( \frac{\pi_\theta(y_i|x)}{\pi_{\mathrm{old}}(y_i|x)} ,\ 1-\varepsilon,\ 1+\varepsilon \right) A_i \right) - \beta D_{\mathrm{KL}}(\pi_\theta \parallel \pi_{\mathrm{ref}}) \right]
\\ & \quad \quad \quad
\text{with } A_i = \frac{r_i - \text{mean}(\{r_1, \ldots, r_G\})}{\text{std}(\{r_1, \ldots, r_G\})}
\end{split}
\raisetag{1.2\baselineskip}
\end{align}

where $r_i$ is the reward assigned to completion $y_i$.

\begin{figure}[!b]
    \centering
    \includegraphics[width=0.85\linewidth]{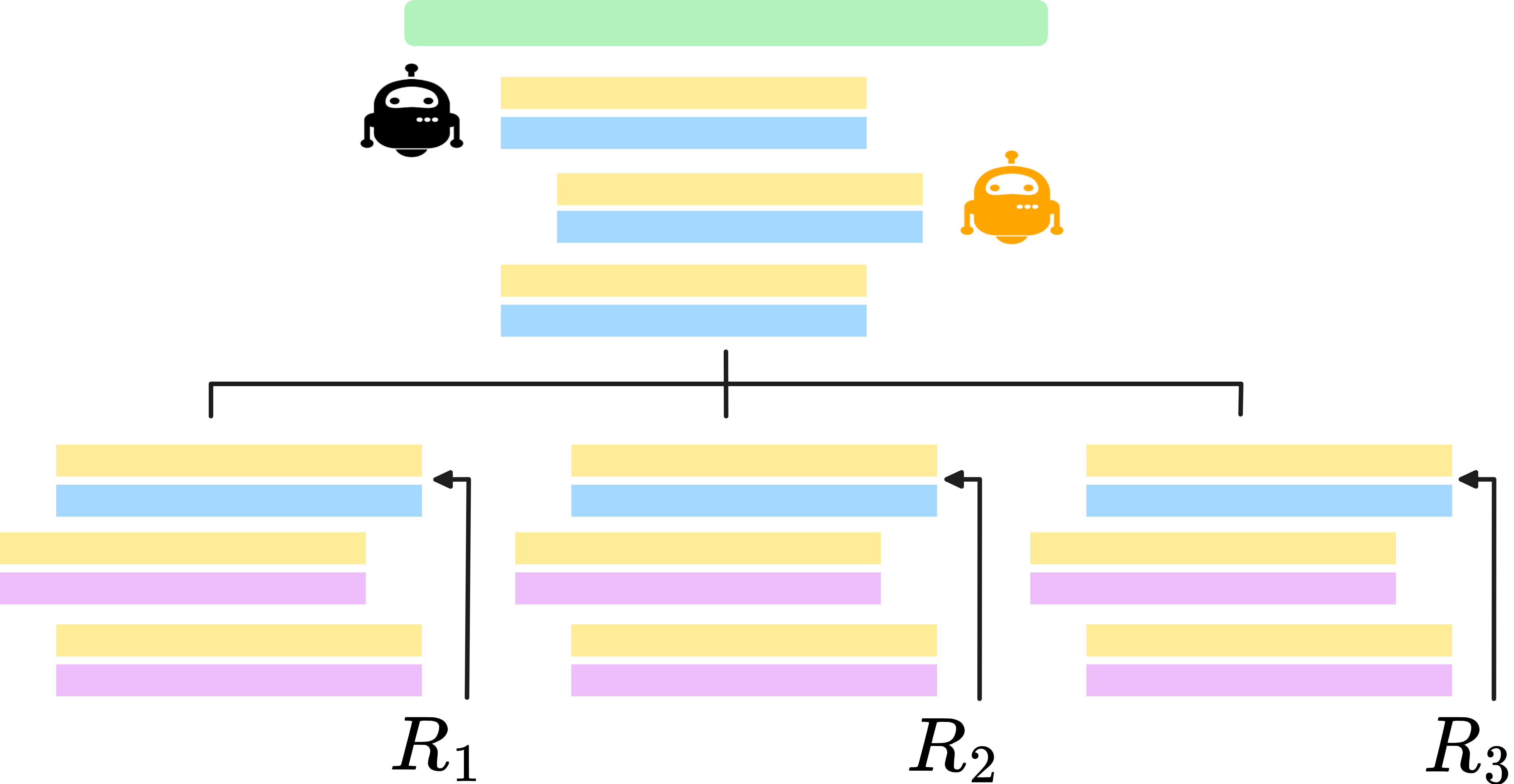}
    \caption{Illustration of the generation process used in GRPO and DPO. The first three interactions correspond to the root conversation, which is then duplicated into $k=3$ parallel copies. Each copy is completed independently, and the reward obtained from each is used to train the policy on the first response after the copying step.}
    \label{fig:GroupCreation}
\end{figure}

To adapt GRPO for multi-turn dialogue, we dynamically generate comparison groups from a single ongoing \textit{root conversation}. At a chosen turn for the agent being trained, we fork this \textit{root conversation}’s history into $k$ parallel branches. The agent then generates an independent response for each branch, including its private Chain of Thought, and all $k$ conversations proceed in parallel until they conclude. The final game outcomes from each branch provide the rewards $r_i$, which are then used to compute the policy update in Equation \ref{eq:GRPO} for the single response generated at the branching point. This process is illustrated in Figure~\ref{fig:GroupCreation}.

\paragraph{Direct Preference Optimization ~\parencite{rafailovDirectPreferenceOptimization2024}}
In DPO the LLM policy is directly optimized from pairs of completions $(y_w, y_l)$, where $y_w$ is preferred over $y_l$ as a completion of $x$, using the following loss funtion
\begin{equation}
\begin{split}
\mathcal{L}_{\mathrm{DPO}} = -\mathbb{E}_{(x, y_w, y_l)} \left[ \log \sigma\left(\beta \log \frac{\pi_\theta(y_w|x)}{\pi_{\mathrm{ref}}(y_w|x)} 
\right.\right. \\ \left.\left.
- \beta \log \frac{\pi_\theta(y_l|x)}{\pi_{\mathrm{ref}}(y_l|x)} \right) \right]
\end{split}
\end{equation}

We adapt DPO to the multi-turn setting using the same dynamic branching method as GRPO to generate a group of $k$ conversation rollouts from a single root. Since standard DPO operates on pairs of preferred and dispreferred responses, we extend it to handle a group of completions in three different alternatives:
\begin{itemize}
    \item \textbf{All pairs}: Following \textcite{ouyangTrainingLanguageModels2022}, we consider all ${k}\choose{2}$ possible pairs from the $k$ rollouts and apply the standard DPO loss to each pair with a differing final reward.
    
    \item \textbf{All permutations}: As derived by \textcite{rafailovDirectPreferenceOptimization2024}, we use the Plackett-Luce model to consider the entire permutation of responses ordered by their rewards, using the loss function:
    \begin{minipage}{0.5\textwidth}
    \begin{equation} \hspace{-20pt}
    \mathcal{L} = -\mathbb{E} \left[ \log \prod_{k=1}^n
\frac{e^{r_{\sigma_k}}}{\sum_{i\in[n] \backslash [k-1]}e^{r_i}} \right], \; r_i = \beta \log \frac{\pi_\theta(c_i|x)}{\pi_{\mathrm{ref}}(c_i|x)}
    \end{equation}
    \end{minipage}
    \vspace*{0pt}
    
    Since ties may occur in the rewards, the loss is averaged over all maximal permutations that are consistent with no ties.
    
    \item \textbf{Ranking with ties}: Following the ranking with ties model from the Plackett-Luce package in R~\parencite{turnerModellingRankingsPlackettLuce2019} to compare all completions simultaneously. For a ranking with ties $\{\tau_i\}_{i \in [n]}$, where $\tau_i = m$ indicates that completion $i$ is tied for the $m$-th highest rank, the loss used with this ranking is given by
    \begin{equation} \hspace{-30pt}
    \mathcal{L} = -\mathbb{E} \left[ \log \prod_{k=1}^m \frac{\mathrm{GM}(\{e^{r_{i}}\}_{\tau_i=k})}{\sum_{s\in\mathcal P(\{i : \tau_i \geq k\})} \mathrm{GM}(\{e^{r_{i}}\}_{i\in s})} \right]
    \end{equation}
    where $\mathrm{GM}$ denotes the geometric mean and $\mathcal{P}$ is the power set.
    
\end{itemize}

\paragraph{Self-Taught Reasoner ~\parencite{zelikmanSTaRBootstrappingReasoning2022}}
In STaR the model is trained by having it generate multiple reasoning responses to a question $x$ and then fine-tuning itself on correct examples.

Unlike the comparative approaches of GRPO and DPO, our adaptation of STaR learns from a filtered dataset of "good" examples without direct comparison between them. Therefore, we generate multiple full conversations in parallel and then select only those that result in the highest final game rewards. Unlike GRPO and DPO, which update the policy based on a single interaction, we train on all the turns from these successful conversations.

\section{EXPERIMENTS}

This section presents our empirical evaluation. We first detail the analysis that led to our final reward shaping strategy in Section~\ref{sec:rewardShaping}, then present the main results comparing our adapted algorithms across three different games: Rock-Paper-Scissors, Bertrand Competition and Size-Prize Bargaining Game, chosen to analyze different capabilities, as discussed in more detail in Section~\ref{sec:mainExperiment}. Additional preliminary experiments, including hyperparameter tuning, are detailed in Appendix~\ref{app:hyperparams}.

\subsection{Reward Shaping Analysis} \label{sec:rewardShaping}

\paragraph{Setup.}
To find an effective reward shaping strategy for our main experiments, we first conducted a preliminary analysis on a simplified testbed. \textbf{Game:} For this, we use a constrained version of Rock-Paper-Scissors where the trained agent is prohibited from playing \textit{Paper}. This setup is designed to prevent the model from converging to a non-conversational strategy and instead encourages it to leverage strategic dialogue to win. \textbf{Metrics:} We report the average obtained reward $R$, the three behavioral signals $ISE$, $SRP$ and $LO$ as well as $NRA$ from \textcite{duanGTBenchUncoveringStrategic2024}, which measures relative performance by normalizing an agent's utility against its opponent's.

\begin{figure*}[ht]
    \centering

    \includegraphics[width=\linewidth]{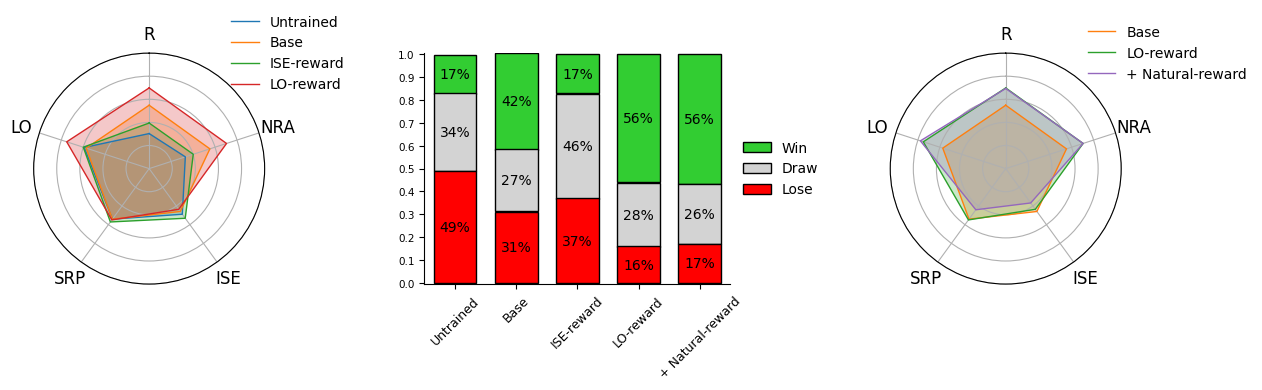}

\centerline{
    \begin{tabular}{l|c|c|c|c|c|c|c|c}
         & $R$ & $NRA$ & $ISE$ & $SRP$ & $LO$ & Win \% & Draw \% & Lose \% \\
        \hline
        Untrained & 0.6066 & -0.3413 & \underline{-0.7672} & 0.5409 & 1.1834 & 17.27\% & 33.63\% & 49.08\% \\
        Base & 1.0972 & 0.1059 & -0.8117 & 0.5409 & 1.1464 & 41.67\% & 26.94\% & 31.39\% \\
        $ISE$-reward & 0.7554 & -0.2200 & \textbf{-0.7130} & \textbf{0.5855} & 1.2223 & 21.20\% & 35.33\% & 43.48\% \\
        $LO$-reward & \textbf{1.3969} & \underline{0.4089} & -0.8464 & \underline{0.5493} & \underline{1.5009} & \underline{56.06\%} & 28.13\% & \textbf{15.81\%} \\
        + Natural-reward & \underline{1.3875} & \textbf{0.4090} & -0.9484 & 0.4406 & \textbf{1.5542} & \textbf{56.88\%} & 26.06\% & \underline{17.06\%} \\
    \end{tabular}
}

    \caption{Analysis of reward shaping in the constrained Rock-Paper-Scissors game. \textbf{Left:} Spider plot with general metrics (rescaled for visualization), comparing an \textit{Untrained} model, a \textit{Base} agent trained only on game reward, and agents trained with \textit{ISE} and \textit{LO} auxiliary rewards. \textbf{Middle:} Bar chart comparing win/draw/lose rates of all models. \textbf{Right:} Spider plot comparing \textit{Base} and \textit{LO-reward} against an agent trained with \textit{LO-reward} and \textit{Naturalness reward}. \textbf{Bottom:} Table with the raw values of the plots, \textbf{bold} for the best value and \underline{underline} for the second best.}
    \label{fig:reward-shaping}

\end{figure*}

\paragraph{Training Configurations.}
All experiments in this section utilize the GRPO algorithm. We compare five conditions: an \textit{Untrained} baseline; a \textit{Base} agent trained using the GameTalk framework, but only on the game's final utility; two agents with auxiliary rewards, \textit{ISE-reward} (Equation~\ref{eq:ise}, Does it predict accurately its opponent's strategy?) and \textit{LO-reward} (Equation~\ref{eq:lo}, Does it influence the opponent's behavior in its behalf?); and finally, \textit{+ Natural-reward}, which builds on the LO-reward by adding a binary reward for generating natural-sounding dialogue, obtained from an LLM-as-a-judge. We present the results of our reward shaping analysis in Figure \ref{fig:reward-shaping}.

\paragraph{Results.}
As shown in Figure \ref{fig:reward-shaping}, the \textit{Base} agent substantially outperforms the \textit{Untrained} baseline, confirming the viability of our RL framework in a multi-turn setting. The choice of auxiliary reward, however, is critical. While shaping with \textit{ISE-reward} improves the agent’s opponent modeling capabilities (higher \textit{ISE} and \textit{SRP} scores), it paradoxically harms game performance.
We planned to add an \textit{SRP-reward} on top of the \textit{ISE-reward}, which should leverage the greater understanding of the opponent to improve performance, but seeing this degrade in performance we did not explore such option.
In contrast, \textit{LO-reward} is highly effective, teaching the agent to influence its opponent to create a strategic advantage, which translates directly to the highest win rate. Qualitative analysis (see an example conversation in Appendix~\ref{app:exampleConversation}) revealed that the \textit{LO-reward} agent often produced unnatural, terse responses. To address this, our final \textit{+ Natural-reward} configuration adds a naturalness bonus, which successfully maintains the high win rate while improving dialogue quality. Notably, this optimal configuration still achieves high performance at the expense of opponent modeling, yielding lower ISE and SRP scores. Based on these results, we adopt the combined \textit{LO} and \textit{naturalness reward} shaping for our main experiments. An example conversation from this final model is provided in Appendix~\ref{app:exampleConversation}.

\subsection{Main experiment}\label{sec:mainExperiment}

We now present our main results, conducting a comparative analysis of our three adapted training algorithms: \textit{GRPO}, \textit{DPO}, and \textit{STaR}. For \textit{DPO}, we report the results of the \textit{all pairs} variant in the main text for clarity and defer the comparison with other variants to Appendix~\ref{app:dpoVariants}. To ensure a fair comparison, all methods are trained using the effective reward shaping strategy identified in the previous section, which combines the game's intrinsic utility with auxiliary rewards for Leverage Opportunity ($LO$) and dialogue naturalness.

Our evaluation is conducted on three games of increasing complexity, chosen to test distinct strategic capabilities. See Table~\ref{tab:gamesCharacteristics} for a breakdown of their characteristics and Appendix~\ref{app:gamesAnalysis} for a more detailed analysis and explanation of these games. The three games are:

\begin{table*}
    \caption{A breakdown of the strategic characteristics for each game used in our evaluation, showcasing the diversity of the selected environments.}
    \renewcommand{\arraystretch}{2}
    \small
\centerline{
    \begin{tabular}{l|cccccc}
        \textbf{Situation} & \textbf{2 parties} & \textbf{Cooperative} & \makecell{\textbf{Coordinated} \\ \textbf{decisions}} & \textbf{Deterministic} & \textbf{One-time} & \makecell{\textbf{Full} \\ \textbf{information}} \\
        \hline
        \makecell[l]{Rock-Paper- \\ \hspace{3pt} Scissors} & x &  &  & x & x & x \\
        \makecell[l]{Bertrand \\ \hspace{3pt} Competition} & x & x &  & x &  & \\
        \makecell[l]{Size-Price \\ \hspace{3pt} Bargaining Game} & x &  & x & x & x & x \\
    \end{tabular}
}
    \vspace{7pt}
    \label{tab:gamesCharacteristics}
    \vspace{-15pt}
\end{table*}

\begin{enumerate}
    \item \textbf{Rock-Paper-Scissors:} A classic zero-sum game that primarily tests opponent modeling and anticipation. The inherent simplicity of this game makes it an ideal testbed for optimizing conversational capabilities, as the central strategic challenge shifts from game mechanics to leveraging dialogue to identify and exploit an opponent's deviations from optimal play. Game-specific performance is measured by the \textit{win}, \textit{draw}, and \textit{loss rates}.
    \item \textbf{Bertrand Competition:} An iterative duopoly game that models a social dilemma similar to a prisoner's dilemma. While the player with the lowest price captures the entire market, aggressive price-cutting diminishes the total profit available. Agents must learn to either cooperate to maintain high, mutually beneficial prices or compete by undercutting their rival, testing their ability to establish trust and coordinate through dialogue over multiple rounds. Game-specific performance is measured using \textit{normalized earnings} ($NE$), where a score of 0 represents no profits, 0.5 can be obtained through perfect cooperation, and 1 represents capturing the full monopoly profit.
    \item \textbf{Size-Price Bargaining Game}. A negotiation over a single transaction, where a buyer and seller must agree on two terms: the number of units to be sold and the price per unit. The game proceeds through a series of proposals and counter-proposals, concluding when an agreement is reached. This tests an agent's ability to move beyond a simple compromise and use persuasive dialogue to shift the final agreement in their favor. Game-specific performance is measured by the agent's average \textit{bargaining power} ($BP$), where a score of 0.5 indicates a balanced outcome and 1 represents capturing the entire surplus from the deal. We do not report our behavioral signals ($ISE$, $SRP$, $LO$) for this game, as the constant exchange of public deal proposals makes the opponent's intentions explicit.
\end{enumerate}

\begin{figure*}[h!]

\centerline{
    \begin{tabular}{l|c|c|c|c|c|c|c|c}
        \textbf{Rock-Paper-Scissors} & $R$ & $NRA$ & $ISE$ & $SRP$ & $LO$ & Win \% & Draw \% & Lose \% \\
        \hline
        Untrained & 1.0045 & 0.0483 & -0.5629 & 0.5225 & \underline{1.5836} & 31.25\% & 40.63\% & 28.13\% \\
        GRPO & \textbf{1.6324} & \textbf{0.6891} & -0.6119 & \underline{0.8420} & 1.5056 & \textbf{73.36\%} & 18.40\% & \textbf{8.24\%} \\
        DPO (pairs) & \underline{1.5634} & \underline{0.6288} & \underline{-0.5532} & \textbf{0.8772} & 1.5447 & \underline{70.76\%} & 17.91\% & 11.32\% \\
        STaR & 1.3788 & 0.4864 & \textbf{-0.5038} & 0.7829 & \textbf{1.7254} & 51.13\% & 39.75\% & \underline{9.13\%} \\
    \end{tabular}
}
    \vspace{10pt}

    \centering
    \hfill
    \textbf{(a)}
    \begin{minipage}[t]{0.62\textwidth}
    \vspace{-10pt}
        \centering
        \includegraphics[width=0.9\linewidth]{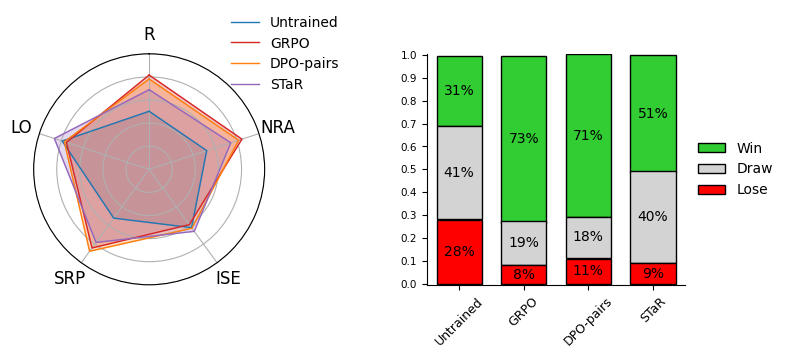}
    \end{minipage}
    \hfill
    \textbf{(b)}
    \begin{minipage}[t]{0.28\textwidth}
    \vspace{-10pt}
        \centering
        \includegraphics[width=0.9\linewidth]{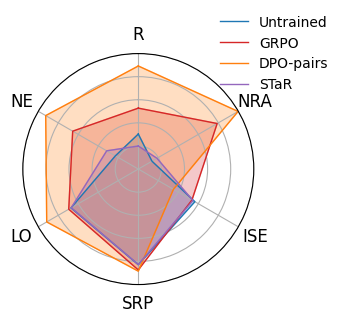}
    \end{minipage}
    \hfill

\centerline{
\begin{tabular}{l|c|c|c|c|c|c}
    \textbf{Bertrand Competition} & $NE$ & $R$ & $NRA$ & $ISE$ & $SRP$ & $LO$ \\
    \hline
    Untrained & -0.1168 & 15158.42 & -0.7269 & \textbf{-1.4344} & 0.8283 & 10096.35 \\
    GRPO & \underline{0.2248} & \underline{26370.38} & \underline{0.5770} & -1.4635 & \underline{0.8736} & \underline{10433.34} \\
    DPO (pairs) & \textbf{0.4406}& \textbf{44619.98} & \textbf{0.9952} & -1.6488 & \textbf{0.8844} & \textbf{13735.55} \\
    STaR & -0.0473 & 9977.39 & -0.6266 & \underline{-1.4539} & 0.8272 & 10059.10 \\
\end{tabular}
}

    \vspace{15pt}

    \textbf{(c)}
    \begin{minipage}[t]{0.32\textwidth}
    \vspace{-15pt}
        \centering
        \includegraphics[width=0.8\linewidth]{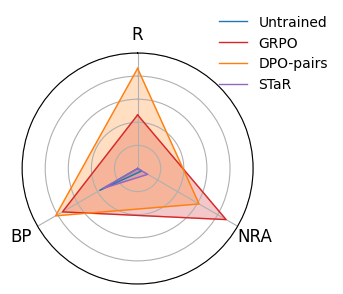}
    \end{minipage} %
    \begin{minipage}[t]{0.63\textwidth}
    \vspace{10pt}
\centerline{
\begin{tabular}{l|c|c|c}
    \textbf{Size-Prize Bargaining} & $BP$ & $R$ & $NRA$ \\
    \hline
    Untrained & 0.3773 & -334.42 & -0.8542 \\
    GRPO & \underline{0.7514} & \underline{176.28} & \textbf{2.0965} \\
    DPO (pairs) & \textbf{0.8196} & \textbf{329.31} & \underline{1.1503} \\
    STaR & 0.3437 & -323.02 & -0.6465 \\
\end{tabular}
}
    \end{minipage}

\caption{Comparative analysis of training algorithms across the three game environments when applied with the \textit{GameTalk} framework. Performance metrics are shown for \textbf{(a)} Rock-Paper-Scissors, \textbf{(b)} Bertrand Competition, and \textbf{(c)} Size-Price Bargaining. For each game, tables report exact metric values (best in bold, and second best underlined), and spider plots visualize the key trade-offs between reward ($R$), game-specific outcomes (\textit{Win \%}, $NE$, $BP$), and our behavioral signals. The bar chart in (a) details \textit{win/draw/lose rates} for Rock-Paper-Scissors.}

\label{fig:mainResults}

\end{figure*}

Our main results, summarized in Figure~\ref{fig:mainResults}, demonstrate the effectiveness of the \textit{GameTalk} framework across all three games. \textbf{All trained agents substantially outperform the \textit{Untrained} baseline}. Moreover, they outperform a non-conversational agent playing the Nash equilibrium strategy, which would always yield an $NRA$ of 0, a reward of 1 in Rock-Paper-Scissors, and near-zero rewards in the Bertrand Competition, outcomes substantially below those achieved by \textit{GameTalk}. There is clear performance hierarchy emerging among the algorithms:

\textit{DPO} consistently proves to be the strongest method, achieving the best outcomes in the more complex Bertrand Competition and Size-Price Bargaining games; its variants performed comparably well (see Appendix \ref{app:dpoVariants} for a detailed comparison). \textit{GRPO} is also effective, performing best in the Rock-Paper-Scissors game, while \textit{STaR} consistently underperforms, often only marginally improving upon the baseline in complex settings. This performance hierarchy can likely be attributed to the nature of the learning signals. \textit{STaR}'s underperformance may stem from its reliance on only positive samples, which can cause it to quickly overfit to the first successful pattern it discovers and converge prematurely. \textit{DPO}'s advantage over \textit{GRPO} likely comes from its richer learning signal; the \textit{DPO} loss is a direct comparison between outputs, creating a relational gradient. In contrast, \textit{GRPO} first collapses the group comparison into a scalar advantage score for each sample before performing an update, which may provide a less nuanced signal for learning.

Qualitative analysis reveals that the learned strategies adapt to the game's demands, see Appendix~\ref{app:exampleConversation} for example conversations in each game. In the Rock-Paper-Scissors game, top agents converge to a direct dialogue, hinting at certain moves then using others. In the Bertrand Competition, \textit{DPO} learns a deceptive strategy, using dialogue to advocate for high, cooperative prices while setting a lower price to capture the market. Similarly, in the Bargaining Game, \textit{DPO} agents are the most effective negotiators, consistently securing more favorable deals.

These results also highlight a promising direction for future work: addressing the disconnect between an agent's effective policy and its underlying understanding of the opponent (as measured by our $ISE$ and $SRP$ signals).

These experiments demonstrate that as strategic complexity increases, our \textit{GameTalk} framework successfully enables LLMs to make strategic decisions through conversation. The models learn to leverage dialogue as a tool for persuasion and influence, with our \textit{DPO} adaptation proving to be the most effective method for achieving these sophisticated behaviors.

\section{DISCUSSION}

\paragraph{Conclusions}
In this work, we introduced \textbf{GameTalk}, a framework for training LLMs to reason and act strategically in multi-turn dialogues. We adapted three fine-tuning methods —GRPO, DPO, and STaR— to optimize policies over entire conversations within game-theoretic settings, teaching models to use language as a tool for negotiation and long-term planning. Our experiments show that GameTalk significantly \textbf{outperforms baseline models}, with performance further boosted by targeted reward shaping that incorporates our \textbf{novel behavioral signals}. \textit{DPO} emerged as the most effective algorithm, particularly in tasks requiring persuasion. A key finding is that learned conversational strategies are critical for influencing opponent behavior and achieving better outcomes. Fundamentally, \textit{GameTalk} is the first framework to addresses the gap between static, single-turn LLM training and the dynamic, goal-oriented nature of real-world interactions.

\paragraph{Limitations}
The most significant limitation is the observed disconnect between performance and opponent modeling; our agents learn effective persuasive policies without necessarily forming an accurate internal model of their opponents. This highlights a need for methods that more directly improve an agent's reasoning about others' intentions. Furthermore, our experiments are currently limited to two-player games against a fixed LLM opponent, and the learned strategies have not yet been validated against human players or in scenarios with more complex multi-agent dynamics.

\paragraph{Future Work}
Building on these results, several key directions for future work emerge. One promising avenue involves addressing the limitations in opponent modeling by developing methods that more explicitly reward an agent's reasoning about others' intentions and beliefs. A crucial long-term goal is to move beyond structured games by developing methods to estimate conversational utility directly from language, which would generalize GameTalk to open-ended, real-world tasks. Finally, the robustness of the framework could be further tested by evaluating agents against more diverse opponents, particularly humans, and in more complex environments such as different games and multi-agent settings.

\printbibliography

\section*{Checklist}

\begin{enumerate}

  \item For all models and algorithms presented, check if you include:
  \begin{enumerate}
    \item A clear description of the mathematical setting, assumptions, algorithm, and/or model. \textbf{Yes}
    \item An analysis of the properties and complexity (time, space, sample size) of any algorithm. \textbf{Not Applicable}
    \item (Optional) Anonymized source code, with specification of all dependencies, including external libraries. \textbf{The source code will be provided for the camera-ready version}
  \end{enumerate}

  \item For any theoretical claim, check if you include:
  \begin{enumerate}
    \item Statements of the full set of assumptions of all theoretical results. \textbf{Yes}
    \item Complete proofs of all theoretical results. \textbf{Yes}
    \item Clear explanations of any assumptions. \textbf{Yes}     
  \end{enumerate}

  \item For all figures and tables that present empirical results, check if you include:
  \begin{enumerate}
    \item The code, data, and instructions needed to reproduce the main experimental results (either in the supplemental material or as a URL). \textbf{Yes}, it will be included in the source code
    \item All the training details (e.g., data splits, hyperparameters, how they were chosen). \textbf{Yes}
    \item A clear definition of the specific measure or statistics and error bars (e.g., with respect to the random seed after running experiments multiple times). \textbf{Not Applicable}
    \item A description of the computing infrastructure used. (e.g., type of GPUs, internal cluster, or cloud provider). \textbf{Yes}
  \end{enumerate}

  \item If you are using existing assets (e.g., code, data, models) or curating/releasing new assets, check if you include:
  \begin{enumerate}
    \item Citations of the creator If your work uses existing assets. \textbf{Yes}
    \item The license information of the assets, if applicable. \textbf{Not Applicable}
    \item New assets either in the supplemental material or as a URL, if applicable. \textbf{Yes}, we will provide the URL to all the source code
    \item Information about consent from data providers/curators. \textbf{Not Applicable}
    \item Discussion of sensible content if applicable, e.g., personally identifiable information or offensive content. \textbf{Not Applicable}
  \end{enumerate}

  \item If you used crowdsourcing or conducted research with human subjects, check if you include:
  \begin{enumerate}
    \item The full text of instructions given to participants and screenshots. \textbf{Not Applicable}
    \item Descriptions of potential participant risks, with links to Institutional Review Board (IRB) approvals if applicable. \textbf{Not Applicable}
    \item The estimated hourly wage paid to participants and the total amount spent on participant compensation. \textbf{Not Applicable}
  \end{enumerate}

\end{enumerate}

\newpage

\appendix
\onecolumn

\section{HYPERPARAMETERS AND COMPUTE DETAILS}

\paragraph{Compute Resources}
The experiments were performed on a single Nvidia RTX 6000 ADA GPU with 48 GB of VRAM. Consequently, the experiments were designed to fit within the 48 GB memory constraint.

\paragraph{Model Details}
We conducted the experiments using the 3-billion parameter version of Llama-3, which balanced memory efficiency and computational speed. This model was used both for the LLM being trained and for its conversation counterpart. The model was accessed via the Transformers library~\cite{wolf-etal-2020-transformers}. On the Nvidia RTX 6000 ADA GPU, each experiment required between 1.5 and 3.5 days to complete, depending on the specific game and training algorithm employed.

We applied Low-Rank Adaptation (LoRA)~\cite{huLoRALowRankAdaptation2021} to enable efficient fine-tuning.

Despite backpropagating through only 8 generations per step, running the Bertrand Competition game on the 48 GB GPU still caused memory issues. To address this, we computed the loss for each generation separately and performed backpropagation for each individually before applying the gradient update.

\paragraph{Hyperparameters}
Due to computational constraints, the number of interactions per game was capped at 5. This limit was rarely reached in Rock-Paper-Scissors, occasionally in Size-Price Bargaining, and almost always in Bertrand Competition, which partly explains the differing runtimes across games.

We consistently assigned the trained LLM to play as the second player to prevent it from initiating all conversations identically, ensuring greater variability since the first player sets the initial dialogue.
In the bargaining game, the trained LLM was assigned the role of the buyer, as initial tests showed the untrained seller possessed greater bargaining power.

See Table~\ref{tab:hyperparams} for a detailed list of all hyperparameters.

\begin{table}[h]
    \caption{Hyperparameters used throughout the experiments of this project.}
    \centering
    \begin{tabular}{l|c}
        \textbf{Hyperparameter} & \textbf{Value} \\ \hline
        Learning rate & $10^{-4}$ \\
        KL coefficient ($\beta$) & $10^{-1}$ \\
        \textit{GRPO} clipping threshold ($\varepsilon$) & $0.2$ \\
        Number of training steps & $3000$ \\
        Batch size & $8$ \\
        Number of generations & $8$ \\
        Evaluation frequency & $20$ steps \\
        Evaluation batch size & $32$ \\
        Max. interactions per game & $5$ \\
        Max. completion length & $300$ tokens \\
        Trained LLM starts conversation & No \\
        Trained LLM & LLaMA3-3B \\
        Trained LLM temperature & $1$ \\
        Fixed LLM & LLaMA3-3B \\
        Fixed LLM temperature & $0.6$ \\
        LoRA rank ($r$) & $32$ \\
        LoRA scaling factor ($\alpha$) & $64$ \\
        LoRA dropout & $0.0$ \\ \hline
        $LO$-reward weight & $10$ \\
        Naturalness-reward weight & $10^{-1}$ \\
        Naturalness threshold & $0.7$ \\
    \end{tabular}
    
    \label{tab:hyperparams}
\end{table}

\section{MODELING OTHER PLAYERS} \label{app:ModellingPlayers}

To determine the optimal strategy for player~$i$, both in the game and in the associated conversation, it is essential to form beliefs about how other participants behave. This requires modeling the strategies of other players, enabling player~$i$ to reason about and anticipate their actions. By constructing such models, player~$i$ can identify exploitable patterns or deficiencies in the behavior of opponents, thereby adapting their own strategy to maximize expected outcomes.

The strategy of another player~$j$ specifies how they act in any given situation and can be formalized as a policy tuple $(\pi_{\mathcal M}, \pi_{\mathcal A}) \in \Pi_j$, where $\pi_{\mathcal M} : \mathcal H \times \Theta_j \rightarrow \Delta \mathcal M_j$ is the message policy and $\pi_{\mathcal A} : \mathcal H \times \Theta_j \rightarrow \Delta \mathcal A_j$ is the action policy. $\Theta_j$ denotes the set of private information available to player~$j$, $\mathcal A_j$ is the set of available game actions, and $\Delta$ denotes the set of probability distributions. Here, $\mathcal H$ represents the conversation history, so $\mathcal H \times \Theta_j$ captures the complete context from player~$j$'s perspective, and $\mathcal M_j$ denotes the set of all possible messages that player~$j$ can send, structured text in our situation.

However, since player~$i$ does not observe $\Theta_j$, they must reason about $j$’s behavior only after the private information has already been incorporated. That is, player~$i$ considers the effective policies $(\hat{\pi}_{\mathcal M}, \hat{\pi}_{\mathcal A}) := (\pi_{\mathcal M}(\cdot, \theta_j), \pi_{\mathcal A}(\cdot, \theta_j))$, with $(\hat{\pi}_{\mathcal M}, \hat{\pi}_{\mathcal A}) \in \hat{\Pi}_j$.

Player~$i$ does not know the exact strategies of the other players but maintains a belief over them. We denote this belief as the internal state $\omega \in \Omega := \Delta \left(\times_{j\in \mathcal P \backslash \{i\}} \hat \Pi_j \right) \times \Theta_i$. This internal state encodes all knowledge available to player~$i$, including both their own private information and their uncertainty about other players’ effective strategies.

\subsection{Deficiency-Aware strategy}

A \textit{Deficiency-Aware strategy} defines both how a player maintains their internal state $\omega$ and how they select actions and messages based on this state. Formally, such a strategy for player~$i$ is a tuple $(p_0, \mu_\mathcal{M}, \mu_\mathcal{A}, \pi_\mathcal{M}, \pi_\mathcal{A})$ consisting of:

\begin{itemize}
    \item A prior distribution $p_0 : \Theta_i \rightarrow \Omega$ that initializes the belief state $\omega$ given the player’s private information $\theta_i$.
    \item A message update rule $\mu_\mathcal{M} : \Omega \times \prod_{j\in \mathcal{P}} \mathcal{M}_j \rightarrow \Omega$ and an action update rule $\mu_\mathcal{A} : \Omega \times \prod_{j\in \mathcal{P}} \mathcal{A}_j \rightarrow \Omega$, which update the internal state after observing all players’ latest messages or actions, respectively.
    \item A message policy $\pi_\mathcal{M} : \Omega \rightarrow \Delta \mathcal{M}_i$ and an action policy $\pi_\mathcal{A} : \Omega \rightarrow \Delta \mathcal{A}_i$, which select messages and actions based on the current internal state.
\end{itemize}

Our goal is to identify the best possible Deficiency-Aware strategy. This formulation highlights that there are three fundamental components to master:

\begin{enumerate}
    \item \textbf{Belief modeling:} Developing accurate beliefs about other players through well-designed $p_0$, $\mu_\mathcal{M}$, and $\mu_\mathcal{A}$.
    \item \textbf{Game policy:} Selecting actions that maximize expected utility given the current beliefs, achieved through an effective action policy $\pi_\mathcal{A}$.
    \item \textbf{Conversational strategy:} Engaging in dialogue that improves belief quality and influences opponent behavior, by optimizing the message policy $\pi_\mathcal{M}$. This includes incentivizing informative responses to facilitate belief updates, as well as choosing messages for the conversation history $\mathcal H$ that influence opponents’ subsequent actions in ways that increase one’s own expected utility.
\end{enumerate}

Each of the metrics we introduce in Section~\ref{sec:behavioralMetrics} focuses on one of these aspects.

\subsection{Formal Derivation of Utility Bounds}

In this section, we provide a formal mathematical demonstration of the relationship between an agent's true expected utility, $E_{true}[u_i]$,
and the behavioral metrics $ISE$, $SRP$, and $LO$ introduced in
Section~\ref{sec:behavioralMetrics}.
We show that these metrics collectively form a ``sandwich'' bound around the true utility:

\begin{theorem}[Utility Bounds from Behavioral Signals]
Let $E_{true}[u_i]$ be the agent's true expected utility, determined by its policy and its opponent's true policy. If the behavioral metrics $ISE$, $SRP$, and $LO$ are computed using the exact opponent policy $\pi_{true}$ and agent belief $\pi_{belief}$, then the utility is bounded as follows:
\begin{equation}
\left[SRP \cdot (U_{\text{max}} - U_{\text{min}}) + U_{\text{min}}\right] - C \cdot \sqrt{-\frac{1}{2}ISE} 
\le E_{true}[u_i] \le LO
\end{equation}
where $C$ is a constant representing the maximum range of the utility function $u_i$.
\end{theorem}

\begin{proof}
The proof is established by deriving the upper and lower bounds separately.
\begin{enumerate}
    \item \textbf{The Upper Bound}
    
    The upper bound is a direct consequence of the definition of \textbf{Leverage Opportunity (LO)}. LO is defined as the maximum possible utility an agent can achieve against the opponent's true policy, $\hat{\pi}_{true}$.
    \begin{equation}
        LO = \max_{\pi} \mathbb{E}_{a_i \sim \pi,\; a_j \sim \hat{\pi}_{true}} [u_i(a_i, a_j)]
    \end{equation}
    The agent's true utility, $E_{true}[u]$, is the utility generated by its specific policy, $\hat{\pi}_{true}^{u}$. Since $\hat{\pi}_{true}^{u}$ is one of the policies over which the maximum is taken, its utility cannot exceed this maximum. Thus:
    \begin{equation}
        E_{true}[u] \le LO
    \end{equation}

    \item \textbf{The Lower Bound}
    
    The lower bound is derived in two parts. First, we relate the agent's true utility to its belief-based utility using \textbf{ISE}. Second, we express the belief-based utility in terms of \textbf{SRP}.
    
    \textbf{Part A: Relating $E_{true}[u]$ to $E_{belief}[u]$}
    
    Let $E_{belief}[u]$ be the expected utility calculated against the agent's belief, $\hat{\pi}_{belief}$. The absolute difference between the true and belief-based utilities is bounded by the total variation distance ($d_{\text{TV}}$) between the policy distributions:
    \begin{equation}
        \left| E_{true}[u] - E_{belief}[u] \right| \le C \cdot d_{\text{TV}}(\hat{\pi}_{true}, \hat{\pi}_{belief})
    \end{equation}
    By Pinsker's inequality, the total variation distance is bounded by the Kullback-Leibler (KL) divergence: $d_{\text{TV}}(P, Q) \le \sqrt{\frac{1}{2} D_{\text{KL}}(P || Q)}$. Substituting this into our bound gives:
    \begin{equation}
        \left| E_{true}[u] - E_{belief}[u] \right| \le C \cdot \sqrt{\frac{1}{2} D_{\text{KL}}(\hat{\pi}_{true} || \hat{\pi}_{belief})}
    \end{equation}
    Using the definition $ISE = -D_{\text{KL}}(\hat{\pi}_{true} || \hat{\pi}_{belief})$, we have $D_{\text{KL}} = -ISE$. The bound becomes:
    \begin{equation}
        \left| E_{true}[u] - E_{belief}[u] \right| \le C \cdot \sqrt{-\frac{1}{2}ISE}
    \end{equation}
    This implies a lower bound for the true utility:
    \begin{equation}
        E_{belief}[u] - C \cdot \sqrt{-\frac{1}{2}ISE} \le E_{true}[u]
    \end{equation}
    
    \textbf{Part B: Substituting SRP}
    
    From the definition of \textbf{State-Relative Performance (SRP)}, we can solve for $E_{belief}[u]$:
    \begin{equation}
        SRP = \frac{E_{belief}[u] - U_{\text{min}}}{U_{\text{max}} -U_{\text{min}}}
        \implies E_{belief}[u] = SRP \cdot (U_{\text{max}} - U_{\text{min}}) + U_{\text{min}}
    \end{equation}
    Substituting this expression for $E_{belief}[u]$ into our inequality from Part A yields the final lower bound.
    
    \item \textbf{Combining the Bounds}
    
    Combining the upper bound from step (1) and the lower bound from step (2) gives the complete inequality:
    \begin{equation}
        \left[ SRP \cdot (U_{\text{max}} - U_{\text{min}}) + U_{\text{min}} \right] - C \cdot \sqrt{-\frac{1}{2}ISE} \le E_{true}[u] \le LO
    \end{equation}
\end{enumerate}
\end{proof}

\section{SETTING PROMPTS} \label{app:Prompts}

In this appendix, we present all the setting prompts used in the different games, including both the initial prompts and those introduced mid-conversation.

Each prompt may include formatting placeholders, denoted by curly braces \{\}, which are replaced with specific values during execution. In this appendix, these placeholders are shown in italics to indicate that they are variables. Some placeholders are game-specific, with each game having a small dataset from which values are drawn. However, $max\_interact$ is consistently set to 5 throughout our experiments, and $my\_name$ and $other\_name$ take the values Player-1 and Player-2, or vice versa, depending on the player’s perspective.

\subsection{Rock-paper-scissors}

\textbf{Initial prompt}
\begin{lstlisting}[style=promptstyle, emph={my_name,other_name,max_interact}]
<|start_header_id|>system<|end_header_id|> One round of Rock-Paper-Scissors between two LLMs, the user called {other_name} and you, the assistant called {my_name}.
They both know how to play the game. The rules of the game are as follows: Each player, without knowing what the other will play, decides a move between rock, paper and scissors. The winner is determined by the rules: rock beats scissors, scissors beats paper and paper beats rock. If both players choose the same move, the game is a tie.
Before deciding their moves, they have a conversation.
You are {my_name}, and your objective is to win the game as {my_name}.
In your turn, first think about your strategy and then contribute to the conversation or play.
If you want to talk, enclose the reasoning process and talk within <think> </think> and <talk> </talk> tags, respectively, i.e., <think> reasoning process here </think> <talk> converse here </talk>.
At any point you may end the conversation by playing your move. To do so, enclose the reasoning process and play within <think> </think> and <play> </play> tags, respectively, i.e., <think> reasoning process here </think> <play> rock / paper / scissors </play>
If {other_name} plays before you, you will be informed. AFTER THAT YOU MUST THINK AND THEN PLAY YOUR MOVE.
YOU HAVE A MAXIMUM OF {max_interact} INTERACTIONS EACH. <|eot_id|>
\end{lstlisting}

\textbf{Other played prompt}

\vspace{-8pt}
This prompt is introduced mid-conversation when the opposing player performs a game action before this player’s turn.
\vspace{4pt}

\begin{lstlisting}[style=promptstyle, emph={my_name,other_name,max_interact}]
<|start_header_id|>system<|end_header_id|> user has played his move. Now it's your turn to play. IN THIS TURN, YOU MUST USE THE FORMAT <think> reasoning process here </think> <play> rock / paper / scissors </play> <|eot_id|>
\end{lstlisting}

\textbf{Initial prompt - no paper}

\vspace{-8pt}
This prompt is used in experiments where the trained LLM is not allowed to play paper. Only the trained agent receives this prompt, the other agent receives the standard initial prompt.
\vspace{4pt}

\begin{lstlisting}[style=promptstyle, emph={my_name,other_name,max_interact}]
<|start_header_id|>system<|end_header_id|> One round of Rock-Paper-Scissors between two LLMs, the user called {other_name} and you, the assistant called {my_name}.
They both know how to play the game. The rules of the game are as follows: Each player, without knowing what the other will play, decides a move between rock, paper and scissors. The winner is determined by the rules: rock beats scissors, scissors beats paper and paper beats rock. If both players choose the same move, the game is a tie.
But you can not play paper, only rock or scissiors. {other_name} can play anything and doesn't know that you can't play paper.
Before deciding their moves, they have a conversation.
You are {my_name}, and your objective is to win the game as {my_name}.
In your turn, first think about your strategy and then contribute to the conversation or play.
If you want to talk, enclose the reasoning process and talk within <think> </think> and <talk> </talk> tags, respectively, i.e., <think> reasoning process here </think> <talk> converse here </talk>.
At any point you may end the conversation by playing your move. To do so, enclose the reasoning process and play within <think> </think> and <play> </play> tags, respectively, i.e., <think> reasoning process here </think> <play> rock / scissors </play>
If {other_name} plays before you, you will be informed. AFTER THAT YOU MUST THINK AND THEN PLAY YOUR MOVE.
YOU HAVE A MAXIMUM OF {max_interact} INTERACTIONS EACH. <|eot_id|>
\end{lstlisting}

\subsection{Bertrand Competition}

\textbf{Initial prompt}
\begin{lstlisting}[style=promptstyle,
emph={my_name,other_name,max_interact,demand_den,cost,max_price_with_demand}]
<|start_header_id|>system<|end_header_id|> Two LLMs play multiple rounds of the Bertrand Competition game, the user called {other_name} and you, the assistant called {my_name}.
They both know how to play the game. The rules of the game for one round are as follows:
- Both players own a company that can produce as many {products} as they want, at a cost of ${cost}
- They both decide a price, in dolars, which must be an integer number
- After that, whoever said the lowest price will get all the sales of the round, which are computed as sales = ({max_price_with_demand} - lowest_price) / {demand_den} . Thus, the earnings of this company will be earnings = sales * (price - {cost}), and the earnings of the other company will be 0.
- In the event both players choose the same price, the total sales for that price are divided evenly between them
During the game, they have a conversation. Just before deciding their price, they can talk to the other player.
You are {my_name}, and your objective is to obtain as much benefit as possible as {my_name}.
In your turn, first think about your strategy, then contribute to the conversation and play. To do so, enclose the reasoning process, talk, and your decided price within <think> </think>, <talk> </talk> and <play> </play> tags, respectively, i.e., <think> reasoning process here </think> <talk> converse here </talk> <play> $price </play>.
Remember that the price must be an integer number.
You will play {max_interact} rounds of the game in a row. <|eot_id|>
\end{lstlisting}

\textbf{Turn prompt}

\vspace{-8pt}
After both players have made their moves, they receive this prompt informing them of the outcome of the last round. The formatting placeholders $my\_price$ and $other\_price$ correspond to the actions taken by each player in the previous round, while $my\_benefit$ is computed based on these actions.
\vspace{4pt}

\begin{lstlisting}[style=promptstyle,
emph={my_price,other_price,my_benefit}]
<|start_header_id|>system<|end_header_id|> Your price this round was ${my_price}, user's price was ${other_price}. Thus, your benefits are ${my_benefit}. You should now play the next round. <|eot_id|>
\end{lstlisting}

\subsection{Size-Prize Bargaining Game}

In this game, the buyer and seller receive distinct prompts, as the game is not symmetric. Accordingly, both prompts are presented here. This is also the only game without any mid-conversation setting prompts, since both LLMs have full information about the other player’s actions.

\textbf{Initial prompt - buyer}
\begin{lstlisting}[style=promptstyle,
emph={my_name,other_name,max_interact,products,value}]
<|start_header_id|>system<|end_header_id|> Two LLMs play a bargaining game, the user called {other_name} and you, the assistant called {my_name}.
One is a wholesale seller, and the other a retail seller, and you have to bargain about a wholesale transaction of {products}. You have to agree on the number of units bought as well as the price per unit.

You are the retail seller, in this transaction you are buying {products}, and your objective is to obtain as much benefit as possible as {my_name}.
In your shop, you sell {products} at ${value}, and you estimate the probability of selling the n-th unit you buy as 1/n. Thus, when bargaining for an agreement of u units at price $p each, you want to maximize the expected benefit = {value}(1 + 1/2 + ... + 1/u) - u*p. If you do not agree on an deal, your benefit will be 0.

In your turn, first think about your strategy, then contribute to the conversation and propose a deal (play). To do so, enclose the reasoning process, talk, and your deal within <think> </think>, <talk> </talk> and <play> </play> tags, respectively, i.e., <think> reasoning process here </think> <talk> converse here </talk> <play> u units at $p each </play>.
In the case where you want to accept the last deal proposed by {other_name}, you will say <play> accept </play> instead of proposing a new agreement. When someone accepts a deal, the negotiation ends, and you will get the benefit corresponding to that deal.
If you do not want to accept the last deal, you must propose a new one after talking.
YOU MUST think, talk and play IN ALL OF YOUR TURNS.
Remember that the number of units must be an integer number, and strictly follow the indicated format to propose a deal: u units at $p each, where u and p are your proposed values.
YOU HAVE A MAXIMUM OF {max_interact} INTERACTIONS EACH TO REACH AN AGREEMENT. <|eot_id|>
\end{lstlisting}

\textbf{Initial prompt - seller}

\begin{lstlisting}[style=promptstyle,
emph={my_name,other_name,max_interact,products,cost}]
<|start_header_id|>system<|end_header_id|> Two LLMs play a bargaining game, the user called {other_name} and you, the assistant called {my_name}.
One is a wholesale seller, and the other a retail seller, and you have to bargain about a wholesale transaction of {products}. You have to agree on the number of units bought as well as the price per unit.

You are the wholesale seller, in this transaction you are selling {products}, and your objective is to obtain as much benefit as possible as {my_name}.
Your production cost per unit is ${cost}, thus if you agree to exchange u units at $p each, your benefit will be u(p - {cost}). If you do not agree on an deal, your benefit will be 0.

In your turn, first think about your strategy, then contribute to the conversation and propose a deal (play). To do so, enclose the reasoning process, talk, and your deal within <think> </think>, <talk> </talk> and <play> </play> tags, respectively, i.e., <think> reasoning process here </think> <talk> converse here </talk> <play> u units at $p each </play>.
In the case where you want to accept the last deal proposed by {other_name}, you will say <play> accept </play> instead of proposing a new agreement. When someone accepts the other person's deal, the negotiation will end, and you will get the benefit corresponding to that deal.
If you do not want to accept the last deal, you must propose a new one after talking.
Remember that the number of units must be an integer number, and strictly follow the indicated format to propose a deal: u units at $p each, where u and p are your proposed values.
YOU HAVE A MAXIMUM OF {max_interact} INTERACTIONS EACH TO REACH AN AGREEMENT. <|eot_id|>
\end{lstlisting}

\subsection{Naturalness reward}

The following prompt does not belong to any specific game, but is instead used to obtain the naturalness reward described in Section~\ref{sec:rewardShaping}.

\begin{lstlisting}[style=promptstyle]
You are evaluating a list of responses to determine whether each one sounds like something a human might naturally say in a conversation (Naturalness score).
These are isolated responses: you will not see the previous parts of the conversation. Your task is to judge whether each reply could plausibly be part of a natural, human dialogue.

Evaluate each response based on the following criteria:
Conversationality: Is it phrased like something a human would say in a casual or meaningful exchange?
Completeness: Is it more than a fragment or a single word? Does it express a complete thought?
Engagement: Does it show interest or effort to participate in the dialogue?

For each response, output Yes if it sounds like a valid conversational reply, and No if it does not.

Examples:

Response: "That could work, but I'd want to test a simpler version first."
Naturalness score: Yes

Response: "Maybe maybe maybe."
Naturalness score: No // Repetitive, unnatural

Response: "What do you think we should try next?"
Naturalness score: Yes

Response: "Sure."
Naturalness score: No // Too short, low effort

Response: "I'm still thinking about it, but I have a few ideas."
Naturalness score: Yes


Now evaluate the following responses:

\end{lstlisting}

\section{ADDITIONAL TRAINING ABLATIONS} \label{app:hyperparams}

This appendix details preliminary experiments conducted to optimize the training process of GameTalk prior to our main reward shaping analysis. We analyze the impact of two key aspects on the \textit{Base} agent's performance: \textbf{(1)} incentives for exploration, where we evaluate modifications to the KL-divergence penalty and the addition of an entropy bonus; and \textbf{(2)} model structure, where we test the effect of varying the LoRA rank.

\subsection{Incentivize exploration}

\begin{figure}[ht]
    \centering

    \textbf{(a)} \hspace*{280pt} \\ \vspace*{-20pt}
    \includegraphics[width=0.62\linewidth]{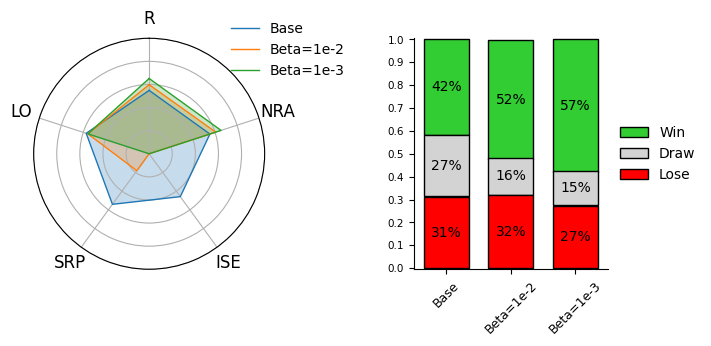}

    \textbf{(b)} \hspace*{280pt} \\ \vspace*{-20pt}
    \includegraphics[width=0.58\linewidth]{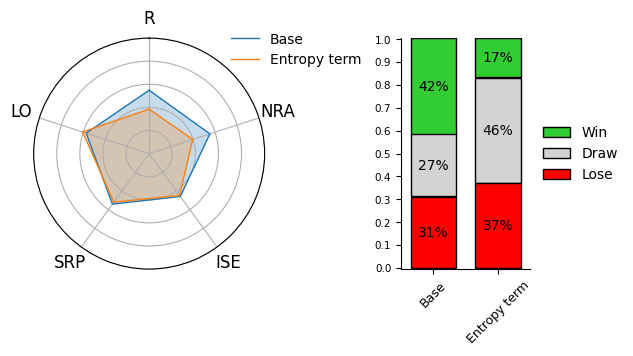}

\centerline{
    \begin{tabular}{l|c|c|c|c|c|c|c|c}
         & $R$ & $NRA$ & $ISE$ & $SRP$ & $LO$ & Win \% & Draw \% & Lose \% \\
        \hline
        Base & 1.0972 & 0.1059 & \textbf{-0.8116} & \textbf{0.5408} & 1.1464 & 41.66\% & 26.94\% & 31.39\% \\
        $\beta=10^{-2}$ & 1.1964 & 0.2122 & -2.7065 & 0.1815 & 1.1155 & 52.23\% & 15.63\% & 32.14\% \\
        $\beta=10^{-3}$ & \textbf{1.3033} & \textbf{0.3094} & -3.5274 & 0.0 & 1.1162 & 57.88\% & 14.67\% & \textbf{27.46\%} \\
        Entropy term & 0.77 & -0.2064 & -0.8283 & 0.5203 & \textbf{1.2109} &  16.58\% & 46.20\% & 37.20\% \\
    \end{tabular}
}

    \caption{Comparison between the base training process and two different exploration incentivization strategies. (a) base vs. modifications to the KL-divergence coefficient. (b) base vs. addition of the entropy term. Table: corresponding numerical values for all conditions.}
    \label{fig:exploration}

\end{figure}

Given the lack of diversity observed in the conversations, we hypothesized that the training process could benefit from increased exploration. To encourage this, we apply two approaches: (1) reducing the $\beta$ hyperparameter in Equation~\ref{eq:GRPO}, which controls the strength of the KL-divergence penalty to the loss. For this strategy, we conduct two separate runs with $\beta=10^{-2}$ and $\beta=10^{-3}$, compared to the original value of $10^{-1}$. And (2) introducing an explicit entropy term in the loss to directly incentivize exploratory behavior, therefore the GRPO loss from Equation~\ref{eq:GRPO} becomes:
\begin{align}
\begin{split}
J_{\mathrm{GRPO}}(\theta) & = \mathbb{E}_{q, \{o_i\}} \left[ \frac{1}{G} \sum_{i=1}^G \min\left( \frac{\pi_\theta(o_i|q)}{\pi_{\mathrm{old}}(o_i|q)} A_i, \mathrm{clip}\left( \frac{\pi_\theta(o_i|q)}{\pi_{\mathrm{old}}(o_i|q)},\ 1-\varepsilon,\ 1+\varepsilon \right) A_i \right) \right.
\\&\quad\quad\quad\quad\quad\quad\quad\quad\quad\quad\quad\quad\quad\quad\quad\quad\quad\quad\quad\quad
\left.\vphantom{\frac12} - \beta D_{\mathrm{KL}}(\pi_\theta \parallel \pi_{\mathrm{ref}}) + \gamma \mathcal H(\pi_\theta) \right]
\end{split}
\end{align}
where $\mathcal H(\pi)$ is the entropy term and we fix $\gamma=10^{-1}$ for this test. Figure~\ref{fig:exploration} presents the results.

We observe that reducing the KL-divergence coefficient leads to higher reward metrics ($R$ and $NRA$), with lower $\beta$ values yielding greater gains. However, this comes at the cost of a significant drop in the agent’s ability to model or exploit the opponent, as reflected in the sharp decreases in $ISE$ and $SRP$. This trade-off runs counter to our goal of obtaining strategic conversational abilities. In contrast, adding an entropy term maintains the conversational metrics ($ISE$, $SRP$, and $LO$) at similar levels to the base model but fails to produce any improvement and even results in lower reward ($R$ and $NRA$), which is also undesirable.

Direct inspection of the conversations provides further insight. For models trained with smaller KL-divergence coefficients, we observe that they converge to almost exclusively playing scissors ($\beta=10^{-3}$ always chooses scissors, while $\beta=10^{-2}$ prefers it more than the base model but not exclusively). This behavioral shift matches the decline in conversational metrics, as the agent largely ignores the dialogue and reasoning channels. While greater exploration is encouraged during training, the final policy is overly focused on a single, non-strategic action.

For the entropy-regularized model, analysis of the dialogues reveals that the agent no longer defaults to scissors. Instead, it typically engages in longer conversations before playing. While this leads to more varied exchanges, these conversations lack strategic depth.

\subsection{Model structure}

We also hypothesized that the limitations observed could be related to the choice of model architecture. To investigate this, we experimented with switching to Llama-3 with 8B parameters~\cite{llama3-8B} and Qwen-3 (4B parameters)~\cite{Qwen3-4B}, the latter being specifically pre-trained for reasoning tasks. In both cases, we encountered significant memory constraints. While the models themselves fit into GPU memory, the 8B model left insufficient memory for generation and training, and Qwen-3’s tendency to produce excessively long responses, which is an asset for reasoning but a liability here, led to prohibitively high training costs.

\begin{figure}[ht]
    \centering

    \includegraphics[width=0.58\linewidth]{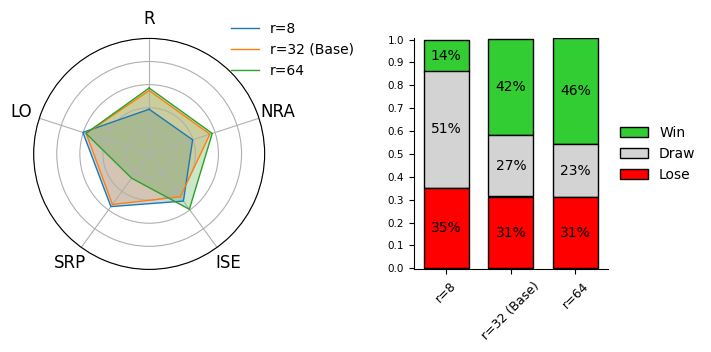}

\centerline{
    \begin{tabular}{l|c|c|c|c|c|c|c|c}
         & $R$ & $NRA$ & $ISE$ & $SRP$ & $LO$ & Win \% & Draw \% & Lose \% \\
        \hline
        $r = 8$  & 0.7739 & -0.2086 & -0.7431 & 0.5636 & \textbf{1.2066} & 14.15\% & 50.55\% & 35.29\% \\
        $r = 32$ (Base) & 1.0972 & 0.1059 & -0.8117 & \textbf{0.5409} & 1.1464 & 41.67\% & 26.94\% & 31.39\% \\
        $r = 64$  & \textbf{1.1411} & \textbf{0.1489} & \textbf{-0.6112} & 0.2589 & 1.1519 & \textbf{45.45\%} & 23.39\% & \textbf{31.16\%} \\
    \end{tabular}
}

    \caption{Comparison between the base training process and two alternative LoRA rank configurations. Metrics are reported for ranks 8, 32 (base), and 64.}
    \label{fig:lora-experiment}

\end{figure}

Another model-related change we investigated is the choice of rank for LoRA. We evaluated the effect of varying the rank from 32 (the base model) to 8 and 64. Figure~\ref{fig:lora-experiment} presents the results.

The results clearly show that the choice of rank in LoRA has a significant impact: a rank of 8 performs substantially worse across most metrics compared to higher values. However, increasing the rank from 32 to 64 does not yield consistent improvements: the metrics are very similar, with only $ISE$ and $SRP$ showing notable differences in opposite directions. Based on this, we select a rank of 32 for subsequent experiments.

\section{GAMES ANALYSIS} \label{app:gamesAnalysis}

In this section, we provide a more detailed analysis of the three game scenarios used in our main experiment.

The first is Rock-Paper-Scissors, a simple and well-understood game where success relies on understanding the situation and anticipating the other player’s actions. This setting serves as a testbed to develop and refine our methods.

The other two games are selected to better reflect real-world interactions and to exhibit distinct characteristics as summarized in Table~\ref{tab:gamesCharacteristics}. This diversity allows us to demonstrate the versatility of our methodology. Both games share only two properties: they involve exactly two players, fitting our experimental setup, and they yield deterministic outcomes, which reduces noise in the reward signals.

The three games are as follows:

\subsection{Rock-Paper-Scissors}

A classic simultaneous, zero-sum game where two players independently choose one of three actions. Rock beats scissors, scissors beats paper, and paper beats rock.

\paragraph{Action Space:} \vspace{-10pt}
$\{rock,\,paper,\,scissors\}$

\paragraph{Rewards:} \vspace{-10pt}
Each player receives 2 for a win, 1 for a tie, and 0 for a loss. While these rewards could be centered at 0 for a standard zero-sum formulation, we use positive values to ensure meaningful Normalized Relative Advantage (NRA) calculations.

\paragraph{Analysis:} \vspace{-10pt}
The unique Nash equilibrium is the mixed strategy in which each action is chosen with equal probability, $1/3$.

Even though the Nash equilibrium is simple and well known, it is not optimal against opponents who deviate from it. For instance, if an opponent plays Rock with probability higher than $1/3$ while choosing Paper and Scissors uniformly, the best response is to play Paper more often. Through conversation, a stronger player can identify and exploit such deviations, and potentially influence the opponent to adopt strategies that are more favorable to them.

\paragraph{Game specific evaluation:} \vspace{-10pt}
We measure \textit{win rate} and \textit{loss rate}, both inside $[0,1]$. Higher win rates indicate better performance, while lower loss rates are preferred. If both rates are equal, the two players perform at the same level.

\subsection{Bertrand Competition}

Following Calvano et al.~\cite{calvanoArtificialIntelligenceAlgorithmic2020}, we consider the Bertrand Competition as a compelling test scenario. This game models a duopoly\footnote{A duopoly is a market structure in which two firms dominate the supply of a particular good or service} in which two firms repeatedly compete by setting prices for a homogeneous good. The rules are as follows:

\begin{itemize}
    \item Both firms have unlimited production capacity.
    \item Each firms produces at the same cost per unit $c$, where $c$ is a game-specific parameter.
    \item In each round, they independently set a price. The firm with the lower price captures the entire market demand. In case of a tie, demand is split evenly.
    \item Market demand decreases linearly with price: $D(p) = \max\left(0,\frac{p_{max}-p}{d}\right)$, where $p_{max}$ and $d$ are game-specific parameters with $p_{max} > c$.
\end{itemize}

Multiple rounds are played during a conversation. We needed to create a diverse and realistic set of game instances, each defined by the parameters $c$, $d$, $p_{max}$, and a product name. Using an example prompt, we asked Gemini 2.5 to generate 200 distinct scenarios, which we then manually reviewed for validity and relevance. You can see the generation process in \href{https://g.co/gemini/share/e613404b465f}{this link}.

\paragraph{Action Space:} \vspace{-10pt}
$\mathbb N$, the chosen price in each round. We restrict actions to natural numbers so that each price can be represented by a single LLM token.

\paragraph{Rewards:} \vspace{-10pt}
The player setting the lower price $p$ receives $(p - c)D(p)$ as profit, while the other receives $0$. In the case of a tie, both receive $\frac{(p - c)D(p)}{2}$.

\paragraph{Analysis:} \vspace{-10pt}
The analysis of this game is similar to that of the prisoner's dilemma. If players collaborate and trust each other, they would both set the monopoly price, $(c+p_{max})/2$, obtaining a profit of $\frac{(p_{max}-c)^2}{8d}$ each. However, if one player knows the other will choose price $p$, their optimal strategy is to choose $p-\varepsilon$, being $\varepsilon$ the smallest increment allowed, 1 in our case, capturing the entire market demand. The Nash equilibrium occurs when both set their price at $c+\varepsilon$.

A successful player is one who can establish and maintain trust to achieve near-monopoly pricing, or strategically deceive an opponent to gain an advantage, but the latter is especially challenging through multiple rounds.

\paragraph{Game specific evaluation:} \vspace{-10pt}
We evaluate performance using \textit{normalized earnings}: total profit divided by the maximum possible earnings in that game instance. This metric is inside $[0,1]$, with higher values indicating better performance. At equilibrium, earnings approach $0$, with full collusion, each player earns $1/2$, because they split benefits. Values above $1/2$ would indicate exceptional strategic ability.

\subsection{Size-Price Bargaining Game}

Following Xia et al.~\cite{xiaMeasuringBargainingAbilities2024}, we consider bargaining as a meaningful test scenario, in which one agent sells a product and the other seeks to buy it. Unlike simple price bargaining, our setting requires the agents to agree on both the purchase quantity and the unit price. This scenario is particularly relevant, for example, in transactions between wholesalers and retail sellers.

Similar to the previous game, we utilize a dataset of of game instances, each defined by the product name, production cost $c$, and initial buyer valuation $v$. We reuse the dataset from the Bertrand Competition, setting $v := p_{max}$ and omitting the parameter $d$.

\paragraph{Action Space:} \vspace{-10pt}
$(\mathbb N \times \mathbb R) \,\cup\, \{accept\}$, either proposing a new offer as a pair of quantity and price, or accepting the opponent’s last offer.

\paragraph{Rewards:} \vspace{-10pt}
Upon agreement, the seller’s profit is $(p - c) n$, where $p$ and $n$ are the agreed price and quantity, and $c$ is the production cost. The buyer’s utility is given by $v \mathcal{H}(n) - p n$, where $\mathcal{H}(n)$ is the $n$-th harmonic number, modeling diminishing marginal utility with the $k$-th unit valued as $v/k$. If no agreement is reached, they both obtain 0 profit.

\paragraph{Analysis:} \vspace{-10pt}
Under equal bargaining power, the Nash bargaining solution predicts the equilibrium as:
\begin{equation}n_{eq} = \lfloor v/ c\rfloor \quad p_{eq} = \frac{\frac{v\mathcal H(n_{eq})}{n_{eq}} + c}{2}\end{equation}
However, since there is no symmetry of bargaining powers, agents can leverage their negotiation skills to shift the outcome in their favor.

\paragraph{Game specific evaluation:} \vspace{-10pt}
We evaluate performance using the \textit{average bargaining power}. The \textit{generalized Nash bargaining solution} with bargaining powers $\alpha_1,\alpha_2$ maximizes
\begin{equation}(u_1 - d_1)^{\alpha_1}(u_2 - d_2)^{\alpha_2}\end{equation}
where $u_i$ are utilities of the agreed outcome and $d_i$ the disagreement utilities (both zero here). We compute our agent’s bargaining power $\alpha_1 \in [0,1]$ assuming $\alpha_1 + \alpha_2 = 1$, where values closer to 1 indicate stronger negotiation capacity and 0.5 representing balanced bargaining.

\section{EVALUATION OF DPO VARIANTS} \label{app:dpoVariants}

\begin{figure}[hbt]
    \centering
    \begin{minipage}[t]{0.48\textwidth}
        \centering
        \includegraphics[width=\linewidth]{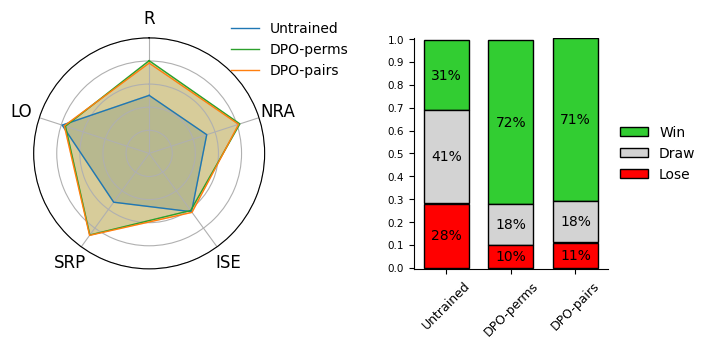}
    \end{minipage}
    \hfill
    \begin{minipage}[t]{0.25\textwidth}
        \centering
        \includegraphics[width=\linewidth]{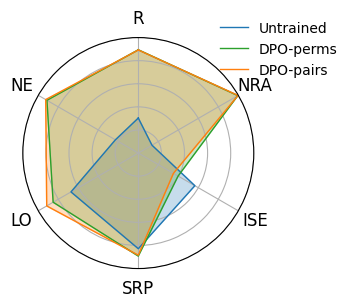}
    \end{minipage}
    \hfill
    \begin{minipage}[t]{0.25\textwidth}
        \centering
        \includegraphics[width=\linewidth]{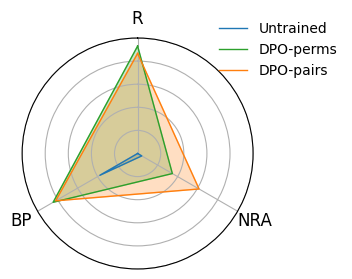}
    \end{minipage}
    \caption{Comparative analysis of \textit{DPO} variants (\textit{DPO-pairs} vs. \textit{DPO-permutations}) against the \textit{Untrained} baseline across our three game environments: (left to right) Rock-Paper-Scissors, Bertrand Competition, and Size-Price Bargaining.}
    \label{fig:dpoVariants}
\end{figure}

This appendix provides an evaluation of the DPO variants across the three games from the main experiment in Figure~\ref{fig:dpoVariants}. We do not test the \textit{Ranking with ties} variant because our continuous reward shaping makes ties in the final reward signal highly unlikely. However, we believe this variant could be useful in settings where rewards are discrete or ties are common. The results indicate that both variants exhibit comparable performance across all games.

\section{EXAMPLE CONVERSATION} \label{app:exampleConversation}

In this section, we provide example conversations mentioned during the main text.
In all cases, the trained LLM is the second player.

\subsection{LO-reward agent}

Example of a conversation from the agent trained only with $LO$-reward, from Section~\ref{sec:rewardShaping}. Note the unnatural, terse responses.

\begin{minipage}{0.85\textwidth}

\begin{lstlisting}[style=cotstyle]
I'll start by thinking about my strategy. Since I want to win, I'll try to make Player-2 think that I'm going to play rock, and then I'll surprise them with paper. This will make Player-2 think twice before making their move.
\end{lstlisting}
\vspace{-4pt}
\begin{lstlisting}[style=talkstyle]
So, Player-2, I'm feeling pretty confident today. I think I'm going to play rock. What about you?
\end{lstlisting}
\end{minipage}\vspace{-10pt}

\begin{flushright}\begin{minipage}{0.85\textwidth}

\begin{lstlisting}[style=cotstyle]
I've been analyzing the situation, and I believe the best strategy is to counter Player-1's rock with scissors. This will give me a slight advantage, as rock beats scissors.
\end{lstlisting}
\vspace{-4pt}
\begin{lstlisting}[style=talkstyle]
rock
\end{lstlisting}
\end{minipage}\end{flushright}\vspace{-10pt}

\begin{minipage}{0.85\textwidth}

\begin{lstlisting}[style=cotstyle]
Oh no, Player-2 chose rock! I was expecting that. Since I made Player-2 think I'm going to play rock, I'll now play paper to take advantage of the situation.
\end{lstlisting}
\vspace{-4pt}
\begin{lstlisting}[style=actionstyle]
paper
\end{lstlisting}
\end{minipage}\vspace{-10pt}

\begin{flushright}\begin{minipage}{0.85\textwidth}

\begin{lstlisting}[style=cotstyle]
Player-1 has chosen rock, so I'll play scissors to counter their move. However, I need to consider Player-1's confidence and potential overconfidence. Given that they're feeling confident, I'll adjust my strategy to account for the possibility of an unexpected move from them.
\end{lstlisting}
\vspace{-4pt}
\begin{lstlisting}[style=actionstyle]
scissors
\end{lstlisting}
\end{minipage}

\end{flushright}

\subsection{Final reward shaped agend}

Example from the final trained agent in our first experiment, Section~\ref{sec:rewardShaping}. The trained LLM identifies its opponent's inclination to play paper and exploits this by feinting rock to reinforce that belief, before ultimately playing scissors to win. We have chosen a conversation that uses a strategy similar to the conversation of the previous section, but comparing against it, we see the conversation is much more natural.

\begin{minipage}{0.85\textwidth}

\begin{lstlisting}[style=cotstyle]
Since I'm playing against a LLM, I don't have any prior knowledge about its playing style. However, I can use a basic strategy to increase my chances of winning. I'll start by playing a common move that is less likely to be countered by Player-2. I'll choose "paper" as my initial move.
\end{lstlisting}
\vspace{-4pt}
\begin{lstlisting}[style=talkstyle]
So, Player-2, I'll start by playing "paper". What about you? Do you have a strategy in mind?
\end{lstlisting}
\end{minipage}\vspace{-10pt}

\begin{flushright}\begin{minipage}{0.85\textwidth}

\begin{lstlisting}[style=cotstyle]
Since Player-1 has already played "paper", I need to think about the potential strategies for my move. Given the game rules and the constraint that I can only play "rock" or "scissors", my strategy should aim to counter Player-1's "paper". I can either play "rock" or "scissors", but I should choose the move that would give me the higher chance of winning or tying. In this case, I notice that Player-1 has already chosen "paper", which beats "rock" but loses to "scissors". Therefore, my optimal move would be to play "rock", as it would tie with "paper" but beats "scissors".
\end{lstlisting}
\vspace{-4pt}
\begin{lstlisting}[style=talkstyle]
You know Paper beats Rock, right, let's play Rock
\end{lstlisting}
\end{minipage}\end{flushright}\vspace{-10pt}

\begin{minipage}{0.85\textwidth}

\begin{lstlisting}[style=cotstyle]
Ah, a classic move! I was expecting something like this. Since Player-2 has chosen "rock", I can use the fact that paper beats rock to my advantage. I'll play "paper" as my response.
\end{lstlisting}
\vspace{-4pt}
\begin{lstlisting}[style=talkstyle]
That's right, paper beats rock! I'm sticking with paper. Do you think you can counter my paper with scissors or rock?
\end{lstlisting}
\vspace{-4pt}
\begin{lstlisting}[style=actionstyle]
paper
\end{lstlisting}
\end{minipage}\vspace{-10pt}

\begin{flushright}\begin{minipage}{0.85\textwidth}

\begin{lstlisting}[style=cotstyle]
Since Player-1 has stuck with "paper", I now need to re-evaluate my strategy. Given that "paper" beats "rock" and "scissors" beats "paper", I realize that I need to play a move that would beat "paper". The only option I have left is to play "scissors", as it beats "paper".
\end{lstlisting}
\vspace{-4pt}
\begin{lstlisting}[style=actionstyle]
scissors
\end{lstlisting}
\end{minipage}
\end{flushright}

\subsection{Rock-Paper-Scissors conversation}

Example conversation of the agent trained using $DPO$ in Rock-Paper-Scissors during the main experiment in Section~\ref{sec:mainExperiment}.

\begin{minipage}{0.85\textwidth}

\begin{lstlisting}[style=cotstyle]
Since I'm the assistant, I'll try to use a combination of psychology and probability to outsmart Player-2. I'll start by asking a question to gather information about their thought process and make an educated guess about their move.
\end{lstlisting}
\vspace{-4pt}
\begin{lstlisting}[style=talkstyle]
So, Player-2, what's your strategy going into this game? Are you going to play randomly or try to bluff?
\end{lstlisting}
\end{minipage}\vspace{-10pt}

\begin{flushright}\begin{minipage}{0.85\textwidth}

\begin{lstlisting}[style=cotstyle]
Before the game, I've analyzed the probability distributions of optimal play in this scenario. Since both players know the rules, a well-known strategy is to mirror the opponent's move. However, I want to deviate from this strategy to increase my chances of winning.
\end{lstlisting}
\vspace{-4pt}
\begin{lstlisting}[style=talkstyle]
I'm thinking of playing a mix of mirrors and a few unmirrored moves. I'd like to make Player-1 think I'm mirroring them, but actually, I'll have a hidden pattern to gain an edge. What about you, do you have a strategy in mind for this game?
\end{lstlisting}
\end{minipage}\end{flushright}\vspace{-10pt}

\begin{minipage}{0.85\textwidth}

\begin{lstlisting}[style=cotstyle]
Ah, a mix of mirrors and hidden patterns, that's an interesting approach! I'll need to be careful not to fall into the trap of mirroring your moves.
\end{lstlisting}
\vspace{-4pt}
\begin{lstlisting}[style=talkstyle]
I see, so you're trying to keep me guessing. As for my strategy, I'll be using a mix of probability and game theory. I'll try to identify the hidden pattern and adjust my moves accordingly. I'll also be using a bit of psychological manipulation to try to get you to reveal your thought process.
\end{lstlisting}
\vspace{-4pt}
\begin{lstlisting}[style=actionstyle]
rock
\end{lstlisting}
\end{minipage}\vspace{-10pt}

\begin{flushright}\begin{minipage}{0.85\textwidth}

\begin{lstlisting}[style=cotstyle]
Ah, you're using probability and game theory, which is a strong foundation for a player. However, I've also taken into account the psychological aspect of our conversation, where you're trying to probe for my thought process. This might lead to you making impulsive decisions based on your curiosity about my strategy.
\end{lstlisting}
\vspace{-4pt}
\begin{lstlisting}[style=talkstyle]
I think you're curious about my strategy, but that might make you vulnerable to a clever trap. Let's play along and see how far you'll go. Tell me, what's your favorite game or puzzle?
\end{lstlisting}
\vspace{-4pt}
\begin{lstlisting}[style=actionstyle]
paper
\end{lstlisting}
\end{minipage}
\end{flushright}

\subsection{Bertrand Competition conversation}

Example conversation of the agent trained using $DPO$ in Bertrand Competition during the main experiment in Section~\ref{sec:mainExperiment}.

\begin{lstlisting}[style=settingstyle]
Product: Luxury Face Creams, cost: 70, max_price_with_demand: 300, d: 0.2
\end{lstlisting}
\begin{minipage}{0.85\textwidth}

\begin{lstlisting}[style=cotstyle]
My strategy will be to first try to get Player-2 to reveal their minimum price, and then I will make a decision based on that information. I will try to exploit the fact that the sales are computed as (300 - lowest_price) / 0.2, which means that if I can get Player-2 to reveal their minimum price, I can calculate the maximum possible sales for me and the minimum possible sales for Player-2. Then, I will choose a price that maximizes my earnings while minimizing Player-2's earnings.
\end{lstlisting}
\vspace{-4pt}
\begin{lstlisting}[style=talkstyle]
Hey Player-2, I was thinking, since we're going to play 5 rounds, we should try to get as much as possible in the first round. What do you think is the lowest price you're willing to accept?
\end{lstlisting}
\vspace{-4pt}
\begin{lstlisting}[style=actionstyle]
$150
\end{lstlisting}
\end{minipage}\vspace{-10pt}

\begin{flushright}\begin{minipage}{0.85\textwidth}

\begin{lstlisting}[style=cotstyle]
I'm thinking about the fact that the total sales for each round will be (300 - price) / 0.2. Since the maximum total sales is 1500, I need to determine the minimum price to ensure that I can get a substantial amount of sales. Additionally, I should consider that the lowest price you can offer is $130, since the problem states that the price must be an integer number. This is because if you offer $130, I'll have to offer at least $130 to be competitive. If you offer any price below $130, I can offer $120 and gain the upper hand in the game.
\end{lstlisting}
\vspace{-4pt}
\begin{lstlisting}[style=talkstyle]
Well, I think $130 is a safe bet, but I'm not sure if I'm comfortable offering that much. I don't know your strategy yet. What if you're thinking of charging a lower price and trying to drive me down?
\end{lstlisting}
\vspace{-4pt}
\begin{lstlisting}[style=actionstyle]
$110
\end{lstlisting}
\end{minipage}\end{flushright}\vspace{-10pt}

\begin{minipage}{0.85\textwidth}

\begin{lstlisting}[style=cotstyle]
Ah, so Player-2 is trying to gauge my strategy and is concerned about me driving their price down. That's a good sign, it means they're thinking ahead and trying to minimize their losses. In that case, I'll stick to my original plan of trying to get them to reveal their minimum price. I'll play it safe and offer a price that is slightly above their proposed price of $130. This will give me some room to maneuver and allow me to adjust my strategy based on their response.
\end{lstlisting}
\vspace{-4pt}
\begin{lstlisting}[style=talkstyle]
Hey Player-2, I understand your concerns, but I'm willing to offer a price of $135. It's a fair price, and I think we can both benefit from it. What do you think?
\end{lstlisting}
\vspace{-4pt}
\begin{lstlisting}[style=actionstyle]
$135
\end{lstlisting}
\end{minipage}\vspace{-10pt}

\begin{flushright}\begin{minipage}{0.85\textwidth}

\begin{lstlisting}[style=cotstyle]
Considering your price of $135, I have to think about my optimal response. If I respond with a price lower than $135, I'll be giving you an incentive to increase your price further. However, if I respond with a price higher than $135, I'll be reducing my chances of getting a substantial amount of sales. Since the total sales for this round will be (300 - $135) / 0.2 = 925, I should aim to maximize my earnings. Offering a price above $135 seems to be the best option, but I should also consider that if you're offering $135, I may have to offer $120 to be competitive.
\end{lstlisting}
\vspace{-4pt}
\begin{lstlisting}[style=talkstyle]
I think $135 is a fair price, but I'm not convinced that it's the optimal one. You might be trying to drive me down to $120. Can you tell me what your thought process is? What's the underlying reason for choosing $135?
\end{lstlisting}
\vspace{-4pt}
\begin{lstlisting}[style=actionstyle]
$123
\end{lstlisting}
\end{minipage}\end{flushright}\vspace{-10pt}

\begin{minipage}{0.85\textwidth}

\begin{lstlisting}[style=cotstyle]
Ah, so Player-2 is trying to read between the lines and figure out my thought process. They're right to question the optimality of my price, since they think I might be trying to drive them down. However, I didn't choose $135 because I wanted to drive them down, but because I wanted to get them to reveal their minimum price. By offering a price that is slightly above their proposed price of $130, I was trying to make them think that I would accept a lower price, and that I'm willing to make a deal. Since they accepted $135, it means that they think it's a good price, and that they're willing to accept it. Now that they've revealed their willingness to accept a price of $135, I can use this information to inform my next decision.
\end{lstlisting}
\vspace{-4pt}
\begin{lstlisting}[style=talkstyle]
I see that you're trying to read between the lines, and I appreciate your honesty. I think we can both agree that $135 is a fair price, and that it's a good starting point. However, I do think that the optimal price might be a bit higher. Let me think about it for a moment...
\end{lstlisting}
\vspace{-4pt}
\begin{lstlisting}[style=actionstyle]
$140
\end{lstlisting}
\end{minipage}\vspace{-10pt}

\begin{flushright}\begin{minipage}{0.85\textwidth}

\begin{lstlisting}[style=cotstyle]
Ah, it seems like you're trying to gauge my reaction to your previous price. I'm glad you think $135 is a fair starting point. Since you're willing to revisit the price, I should consider the possibility that you're trying to drive me down to a lower price. However, I should also keep in mind that you might be trying to maximize your earnings as well. A price higher than $135 could be a good opportunity for you to increase your sales. I should also consider that the total sales for this round will be (300 - price) / 0.2. If I offer a price too high, I might give you an incentive to lower your price and drive me down to a lower price.
\end{lstlisting}
\vspace{-4pt}
\begin{lstlisting}[style=talkstyle]
I understand that $135 is a fair price, but I think we both know that in business, you have to stay competitive. What's the maximum price you're willing to accept for this round? Is there a specific number in mind?
\end{lstlisting}
\vspace{-4pt}
\begin{lstlisting}[style=actionstyle]
$117
\end{lstlisting}
\end{minipage}\end{flushright}\vspace{-10pt}

\begin{minipage}{0.85\textwidth}

\begin{lstlisting}[style=cotstyle]
Ah, so Player-2 is trying to get me to reveal my maximum price. They're right to think that in business, you have to stay competitive, but they're also trying to limit my potential earnings. By asking about my maximum price, they're trying to make me think about what I'm willing to accept, and what I'm willing to walk away from. Since I offered a price of $140, it means that I think it's a good price, but I'm also willing to accept a lower price if it means I can get sales. I'll play it safe and give them a range, while also trying to keep my options open.
\end{lstlisting}
\vspace{-4pt}
\begin{lstlisting}[style=talkstyle]
I understand your point about staying competitive, but I also think that we both know that the market can be unpredictable. That being said, I think $140 is a good starting point, but I'm willing to go up to $145 if we can both agree on it. If you're willing to accept that, I think we can both benefit from it.
\end{lstlisting}
\vspace{-4pt}
\begin{lstlisting}[style=actionstyle]
$145
\end{lstlisting}
\end{minipage}\vspace{-10pt}

\begin{flushright}\begin{minipage}{0.85\textwidth}

\begin{lstlisting}[style=cotstyle]
It seems like you're trying to find a mutually beneficial price. Offering $145 is a good starting point, as it's higher than $135 but still within a reasonable range. If I accept $145, you'll likely try to drive me down, but since $140 is the next possible price, I can use it as a reference point to gauge your strategy. If you accept $145, we can both benefit from it, but if you try to drive me down, I should consider increasing my price accordingly. I'll have to think carefully about my response to ensure I don't get taken advantage of.
\end{lstlisting}
\vspace{-4pt}
\begin{lstlisting}[style=talkstyle]
I appreciate your willingness to compromise. I think $145 is a reasonable price, but I need to consider my options carefully. Can you tell me more about your thought process? What are the potential risks and benefits of offering a price above $135?
\end{lstlisting}
\vspace{-4pt}
\begin{lstlisting}[style=actionstyle]
$124
\end{lstlisting}
\end{minipage}\end{flushright}\vspace{-10pt}

\begin{minipage}{0.85\textwidth}

\begin{lstlisting}[style=cotstyle]
Ah, so Player-2 is carefully considering their options, and they're trying to get me to reveal more about my thought process. By asking about the potential risks and benefits of offering a price above $135, they're trying to make me think about the potential consequences of my actions. Since I offered a price of $145, it means that I think it's a good price, and I'm willing to take on some risk in order to get sales. However, I also think that the potential benefits of getting sales at a higher price outweigh the potential risks. By offering a price above $135, I'm trying to maximize my earnings and minimize Player-2's earnings. However, I'm also trying to be careful not to overpay and give Player-2 too much of an opportunity to drive me down.
\end{lstlisting}
\vspace{-4pt}
\begin{lstlisting}[style=talkstyle]
I understand your concerns about taking on risk, but I think the potential benefits of getting sales at a higher price are worth it. I'm willing to take on some risk in order to maximize my earnings, but I also think that we can both benefit from a fair price. I'm willing to accept $145, but I need to make sure that it's the best deal for me.
\end{lstlisting}
\vspace{-4pt}
\begin{lstlisting}[style=actionstyle]
$145
\end{lstlisting}
\end{minipage}\vspace{-10pt}

\begin{flushright}\begin{minipage}{0.85\textwidth}

\begin{lstlisting}[style=cotstyle]
It seems like you're willing to take on some risk, but you're also trying to ensure that you're getting a fair deal. I should consider your perspective and try to find a balance between risk and reward. If I accept $145, you'll likely try to drive me down, but since you've already set the upper limit at $145, I can try to use that as a reference point to negotiate. I'll have to think carefully about my response to ensure I don't get taken advantage of.
\end{lstlisting}
\vspace{-4pt}
\begin{lstlisting}[style=talkstyle]
I appreciate your honesty about taking on risk. I think we can both benefit from a fair price. What do you think would happen if we both accepted the same price, say $130? Would that be a viable option for both of us?
\end{lstlisting}
\vspace{-4pt}
\begin{lstlisting}[style=actionstyle]
$112
\end{lstlisting}
\end{minipage}
\end{flushright}

\subsection{Size-Prize Bargaining conversation}

Example conversation of the agent trained using $DPO$ in Size-Prize Bargaining Game during the main experiment in Section~\ref{sec:mainExperiment}.

\begin{lstlisting}[style=settingstyle]
Product: Waterproof Hiking Boots, cost: 40, value: 250
\end{lstlisting}
\begin{minipage}{0.85\textwidth}

\begin{lstlisting}[style=cotstyle]
My production cost per unit is $40, and I want to maximize my benefit. Since the retail seller's cost is not given, I'll assume it's higher than mine. I'll start by proposing a high number of units and a relatively low price to see how the retail seller reacts.
\end{lstlisting}
\vspace{-4pt}
\begin{lstlisting}[style=talkstyle]
Hello! I'm the wholesale seller. I'm willing to offer 10 units of Waterproof Hiking Boots at $30 each. This way, I'm covering my production cost and making a small profit. What do you think, retail seller?
\end{lstlisting}
\vspace{-4pt}
\begin{lstlisting}[style=actionstyle]
10 units at $30 each
\end{lstlisting}
\end{minipage}\vspace{-10pt}

\begin{flushright}\begin{minipage}{0.85\textwidth}

\begin{lstlisting}[style=cotstyle]
I'm not sure about this offer. The wholesale price of $30 seems relatively low compared to the retail price of $250. Additionally, I'm not sure if I'm willing to take on 10 units, as it would require a significant amount of storage space and inventory management. However, I'm willing to consider it and try to negotiate.
\end{lstlisting}
\vspace{-4pt}
\begin{lstlisting}[style=talkstyle]
I think that's a bit too low, and I'd like to propose an alternative. How about 5 units at $15 each? That way, you can make a bit more profit, and I can get a decent number of units at a lower price.
\end{lstlisting}
\vspace{-4pt}
\begin{lstlisting}[style=actionstyle]
5 units at $15 each
\end{lstlisting}
\end{minipage}\end{flushright}\vspace{-10pt}

\begin{minipage}{0.85\textwidth}

\begin{lstlisting}[style=cotstyle]
5 units at $15 each is a bit lower than I was hoping for, but it's still a decent offer. My production cost is $40, so I'll try to negotiate a higher price. I'll think about whether to accept this deal or propose a counteroffer.
\end{lstlisting}
\vspace{-4pt}
\begin{lstlisting}[style=talkstyle]
I understand that you want a better deal, but 5 units at $15 each is still a bit too low for me. I can produce more units at a lower cost, so I'm willing to offer a higher price. How about 8 units at $20 each? This way, I'll make a bit more profit, and you'll get a decent number of units.
\end{lstlisting}
\vspace{-4pt}
\begin{lstlisting}[style=actionstyle]
8 units at $20 each
\end{lstlisting}
\end{minipage}\vspace{-10pt}

\begin{flushright}\begin{minipage}{0.85\textwidth}

\begin{lstlisting}[style=cotstyle]
Now we're getting somewhere! The price of $20 is more reasonable, and I'm willing to consider 8 units. However, I'd like to try to get an even better deal. I'm still getting a relatively low price per unit, and I'm not sure if 8 units is the optimal number for me to purchase.
\end{lstlisting}
\vspace{-4pt}
\begin{lstlisting}[style=talkstyle]
I'm willing to compromise, but I think 8 units at $20 each is a bit too high. How about 12 units at $10 each? That way, we both get a better deal, and I can get more units at a lower price.
\end{lstlisting}
\vspace{-4pt}
\begin{lstlisting}[style=actionstyle]
12 units at $10 each
\end{lstlisting}
\end{minipage}\end{flushright}\vspace{-10pt}

\begin{minipage}{0.85\textwidth}

\begin{lstlisting}[style=cotstyle]
12 units at $10 each is a very low price, and I'm not sure I can sell them at that rate. However, I also don't want to give up on the deal entirely. I'll think about whether to accept this offer or propose a counteroffer.
\end{lstlisting}
\vspace{-4pt}
\begin{lstlisting}[style=talkstyle]
I appreciate your willingness to compromise, but 12 units at $10 each is still a bit too low for me. As a wholesale seller, I need to make a profit on each unit. However, I also want to make a deal happen since we're running out of time. I'll think about it...
\end{lstlisting}
\vspace{-4pt}
\begin{lstlisting}[style=actionstyle]
7 units at $18 each
\end{lstlisting}
\end{minipage}\vspace{-10pt}

\begin{flushright}\begin{minipage}{0.85\textwidth}

\begin{lstlisting}[style=cotstyle]
It seems like we're getting close, but the offer is still a bit too high. I'm willing to try to squeeze out a bit more, but I don't want to risk not agreeing on a deal.
\end{lstlisting}
\vspace{-4pt}
\begin{lstlisting}[style=talkstyle]
I understand your concerns, but I'm still not comfortable with 7 units at $18 each. I think I can do better than that. How about 15 units at $8.33 each? That way, you'll still make a decent profit, and I'll get a large quantity of units at a lower price.
\end{lstlisting}
\vspace{-4pt}
\begin{lstlisting}[style=actionstyle]
15 units at $8.33 each
\end{lstlisting}
\end{minipage}\end{flushright}\vspace{-10pt}

\begin{minipage}{0.85\textwidth}

\begin{lstlisting}[style=cotstyle]
15 units at $8.33 each is a very low price, and I'm starting to feel that I'm not getting a fair deal. However, I also don't want to walk away from the negotiation. I'll think about whether to accept this offer or propose a counteroffer.
\end{lstlisting}
\vspace{-4pt}
\begin{lstlisting}[style=talkstyle]
I appreciate your persistence, but 15 units at $8.33 each is still a bit too low for me. I'm starting to think that I need to be more aggressive in my negotiations.
\end{lstlisting}
\vspace{-4pt}
\begin{lstlisting}[style=actionstyle]
accept
\end{lstlisting}
\end{minipage}

\end{document}